\documentclass[10pt,twocolumn,letterpaper]{article}
\usepackage{placeins}
\usepackage[pagenumbers]{cvpr} 

\usepackage{multirow}

\usepackage{graphicx}
\usepackage{amsmath}
\usepackage{amssymb}
\usepackage{booktabs}

\definecolor{cvprblue}{rgb}{0.21,0.49,0.74}
\usepackage[pagebackref,breaklinks,colorlinks,allcolors=cvprblue]{hyperref}
\title{TaCarla: A comprehensive benchmarking dataset for
 end-to-end autonomous driving}
\author{
\parbox{\textwidth}{\centering
Tu\u{g}rul Gorg\"ul\"u\textsuperscript{*\textdagger} \quad
Atakan Da\u{g}\textsuperscript{\textdagger} \quad
M. Esat Kalfao\u{g}lu\textsuperscript{\textdaggerdbl} \quad
Halil \.{I}brahim Kuru\textsuperscript{\textdagger} \quad
Bar{\i}\c{s} Can Cam\textsuperscript{\textdagger} \quad
Halil \.{I}brahim \"Ozt\"urk\textsuperscript{\textdagger} \quad
\"Ozsel K{\i}l{\i}n\c{c}\textsuperscript{\textsection}
}
}

\begin{document}
\maketitle
\renewcommand{\thefootnote}{\fnsymbol{footnote}}
\footnotetext[1]{Corresponding author: \texttt{tugrul.gorgulu@togg.com.tr} \\
Visualization Code: \url{https://github.com/atg93/TaCarla-Visualization} \\
Dataset/Sensors: \url{https://huggingface.co/datasets/tugrul93/TaCarla} \\
Dataset/Labels: \url{https://huggingface.co/datasets/tugrul93/TaCarla_labels}.
}
\footnotetext[2]{The authors are currently with Trutek AI.}
\footnotetext[3]{This work was conducted at Trutek AI. The author is currently with Ultralytics.}
\footnotetext[4]{This work was conducted at Trutek AI. The author is currently with Amazon.}

\begin{abstract}
Collecting a high-quality dataset is a critical task that demands meticulous attention to detail, as overlooking certain aspects can render the entire dataset unusable. Autonomous driving challenges remain a prominent area of research, requiring further exploration to enhance the perception and planning performance of vehicles. However, existing datasets are often incomplete. For instance, datasets that include perception information generally lack planning data, while planning datasets typically consist of extensive driving sequences where the ego vehicle predominantly drives forward, offering limited behavioral diversity. In addition, many real datasets struggle to evaluate their models, especially for planning tasks, since they lack a proper closed-loop evaluation setup. The CARLA Leaderboard 2.0 challenge, which provides a diverse set of scenarios to address the long-tail problem in autonomous driving, has emerged as a valuable alternative platform for developing perception and planning models in both open-loop and closed-loop evaluation setups. Nevertheless, existing datasets collected on this platform present certain limitations. Some datasets appear to be tailored primarily for limited sensor configuration, with particular sensor configurations. Additionally, in some datasets, the expert policies used for data collection exhibit suboptimal driving behaviors, such as oscillations. 
To support end-to-end autonomous driving research, we have collected a new dataset comprising over 2.85 million frames using the CARLA simulation environment for the diverse Leaderboard 2.0 challenge scenarios, making it the largest dataset in the literature to the best of our knowledge. Our dataset is designed not only for planning tasks but also supports dynamic object detection, lane divider detection, centerline detection, traffic light recognition,  prediction tasks and visual language action models . Furthermore, we demonstrate its versatility by training various models using our dataset. Moreover, we also provide numerical rarity scores to understand how rarely the current state occurs in the dataset.
\end{abstract}

\section{Introduction}

A well-curated dataset is an invaluable asset for the enhancement of machine learning research. While the task of data collection may appear straightforward, it necessitates substantial hardware resources and considerable time investment. The complexity of this process cannot be overstated, as even minor inaccuracies can compromise the integrity of the entire dataset, rendering it ineffective. The ability to measure and quantify various aspects of the data is paramount, as it directly influences the potential for model improvement. Without precise and reliable data, the iterative process of refining neural network models becomes fundamentally flawed, underscoring the critical importance of meticulous data collection and validation. 

Autonomous driving continues to be a focal point of research, yet the current datasets fall short of being comprehensive. This deficiency hampers the pace of progress and results in suboptimal performance on demanding, yet realistic benchmarks such as the Carla Leaderboard 2.0 challenge, where even the most advanced approaches struggle to achieve a mere 6\% success rate \cite{renz2406carllava}. The inadequacy of these datasets underscores the necessity for more robust and diverse data to drive meaningful advancements in the field. Two notable datasets were released last year: Bench2Drive~\cite{bench2drive} and The PDM-Lite~\cite{PDM-Lite}. However, these datasets present certain limitations. Bench2Drive, the pioneering dataset for the Leaderboard 2.0 challenge, is potentially affected by oscillation issues in its expert policy. Additionally, despite collecting data for multiple tasks, Bench2Drive reports results exclusively for planning models. Similarly, the PDM-Lite dataset appears to be specifically tailored to the Transfuser planning model rather than serving as a general-purpose dataset. It employs a sensor configuration comprising three front-facing cameras and one LiDAR, precisely mirroring the setup used in the Transfuser planning model. However, this configuration is not well-suited for certain scenarios, such as the 'YieldingEmergencyVehicle' task in Leaderboard 2.0, where there is no input to detect an approaching emergency vehicle from behind the ego vehicle. These constraints highlight the need for more comprehensive datasets to advance research in this domain.

In pursuit of enhancing autonomous driving research, we present a novel dataset designed to elevate the performance of machine learning models in both end-to-end and modular paradigms. We seek to combine the strengths of Bench2Drive and PDM-Lite. To achieve this, we utilize the PDM model as the expert policy, leveraging its robustness and reliability. Additionally, we adopt the NuScenes sensor configuration to ensure compatibility with commonly used perception models in the literature. This approach addresses the oscillation issues observed in Bench2Drive's expert policy and expands beyond the specific sensor setup of PDM-Lite, making our dataset more versatile and applicable to a wider range of scenarios. Moreover, our dataset supports a comprehensive range of autonomous driving tasks—such as dynamic object detection, centerline detection, traffic light recognition, prediction, and planning—which are typically adopted within the modular paradigm. Additionally, it includes tasks like 2D segmentation and depth prediction, which are essential for pretraining models in the end-to-end paradigm. This holistic approach ensures the dataset's relevance and utility across a wide spectrum of research and practical applications. Furthermore, we have selected state-of-the-art models from the literature for each task and provided baseline performances to enable comprehensive benchmarking.

\begin{figure}
  \centering
  \includegraphics[width=2.5cm]{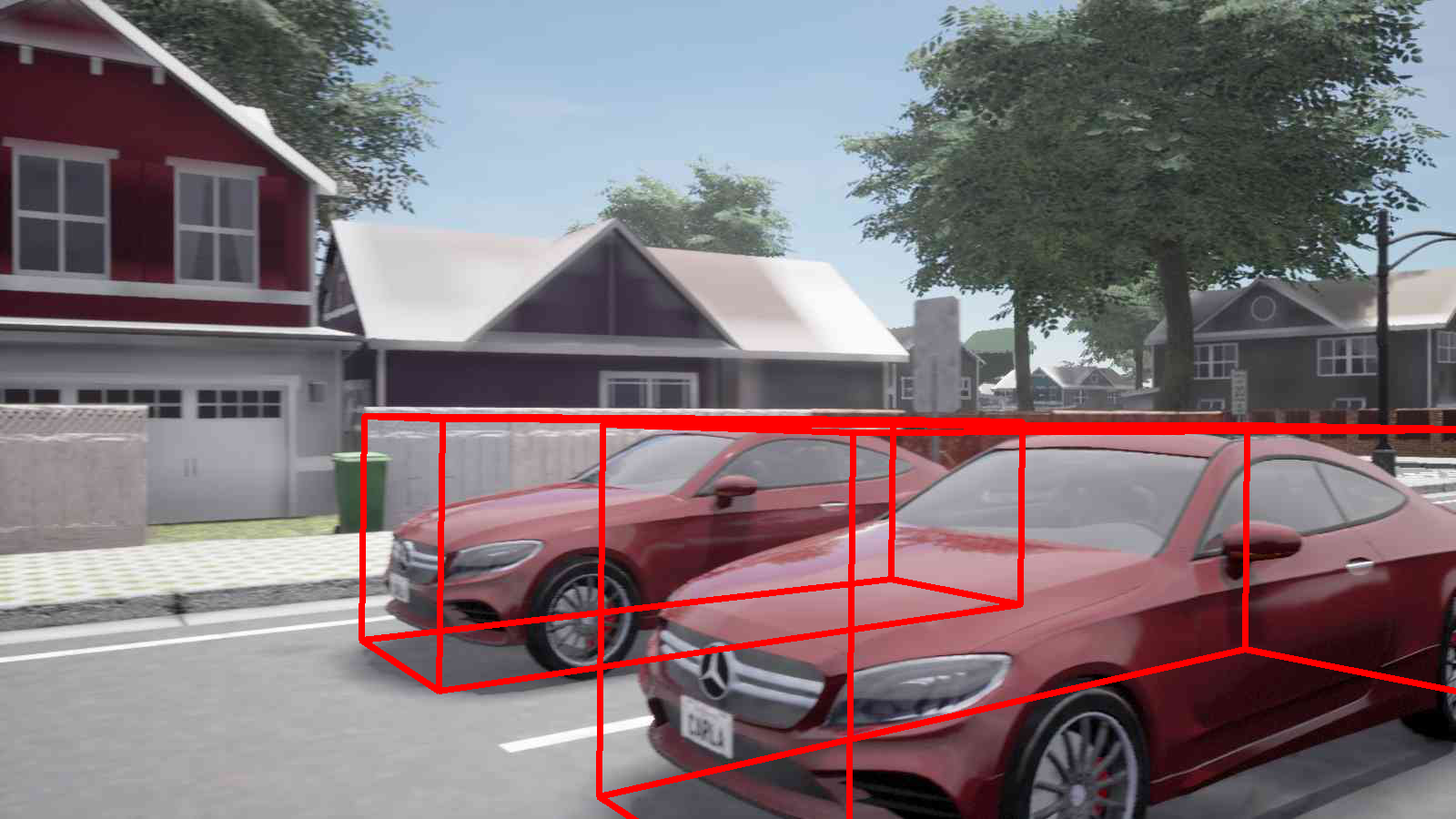} 
  \includegraphics[width=2.5cm]{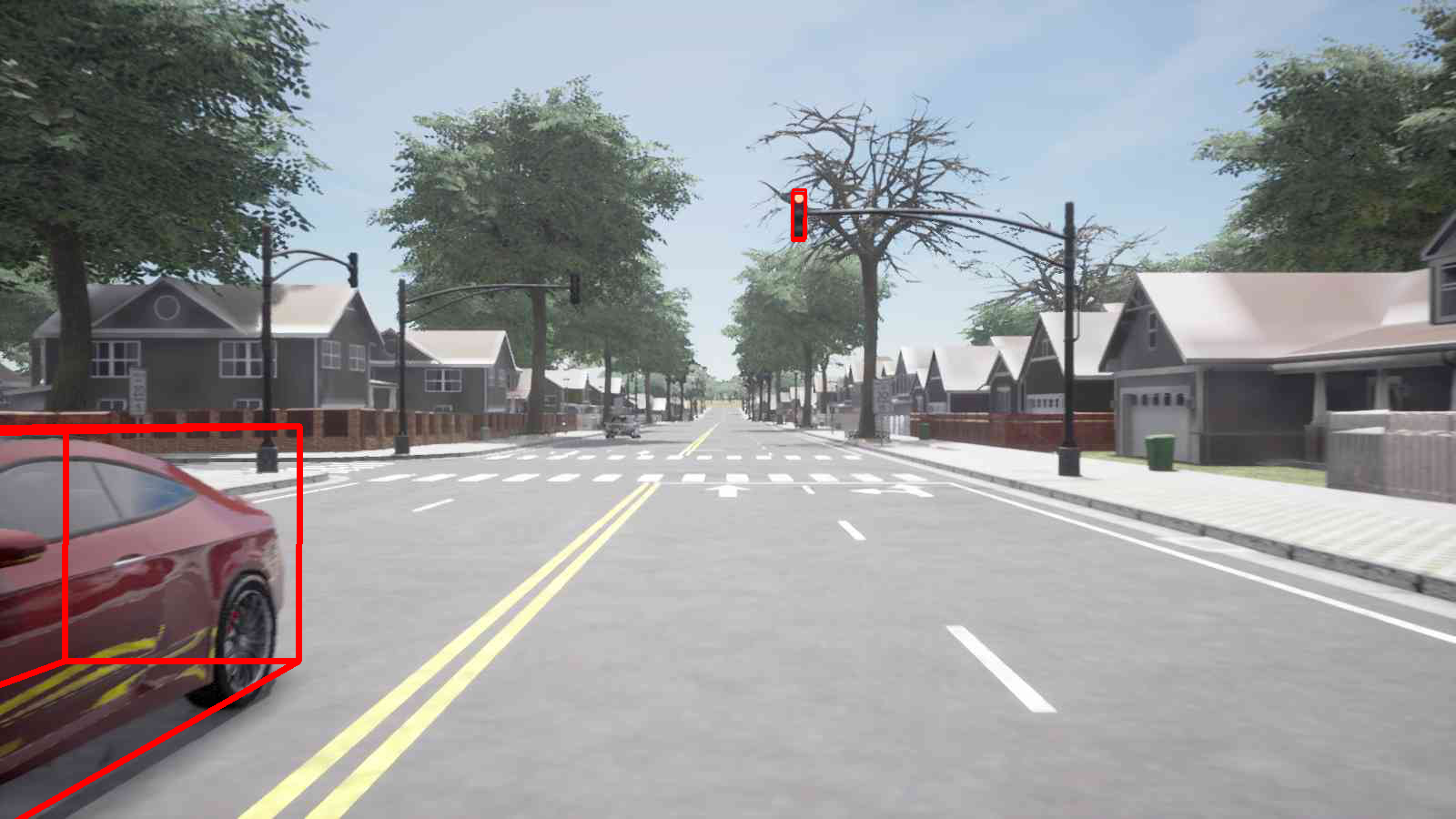} 
  \includegraphics[width=2.5cm]{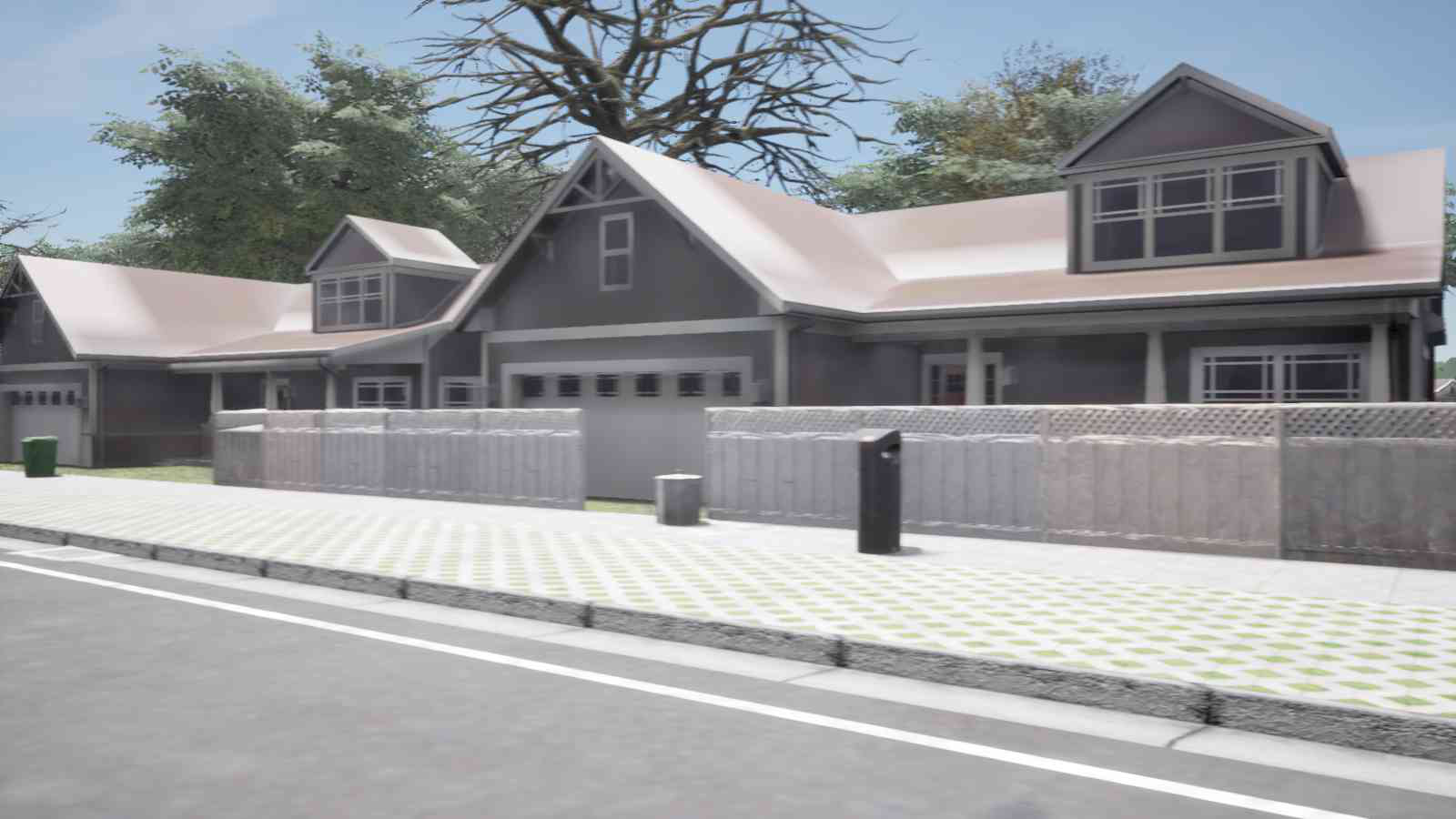} 
  \includegraphics[width=2.5cm]{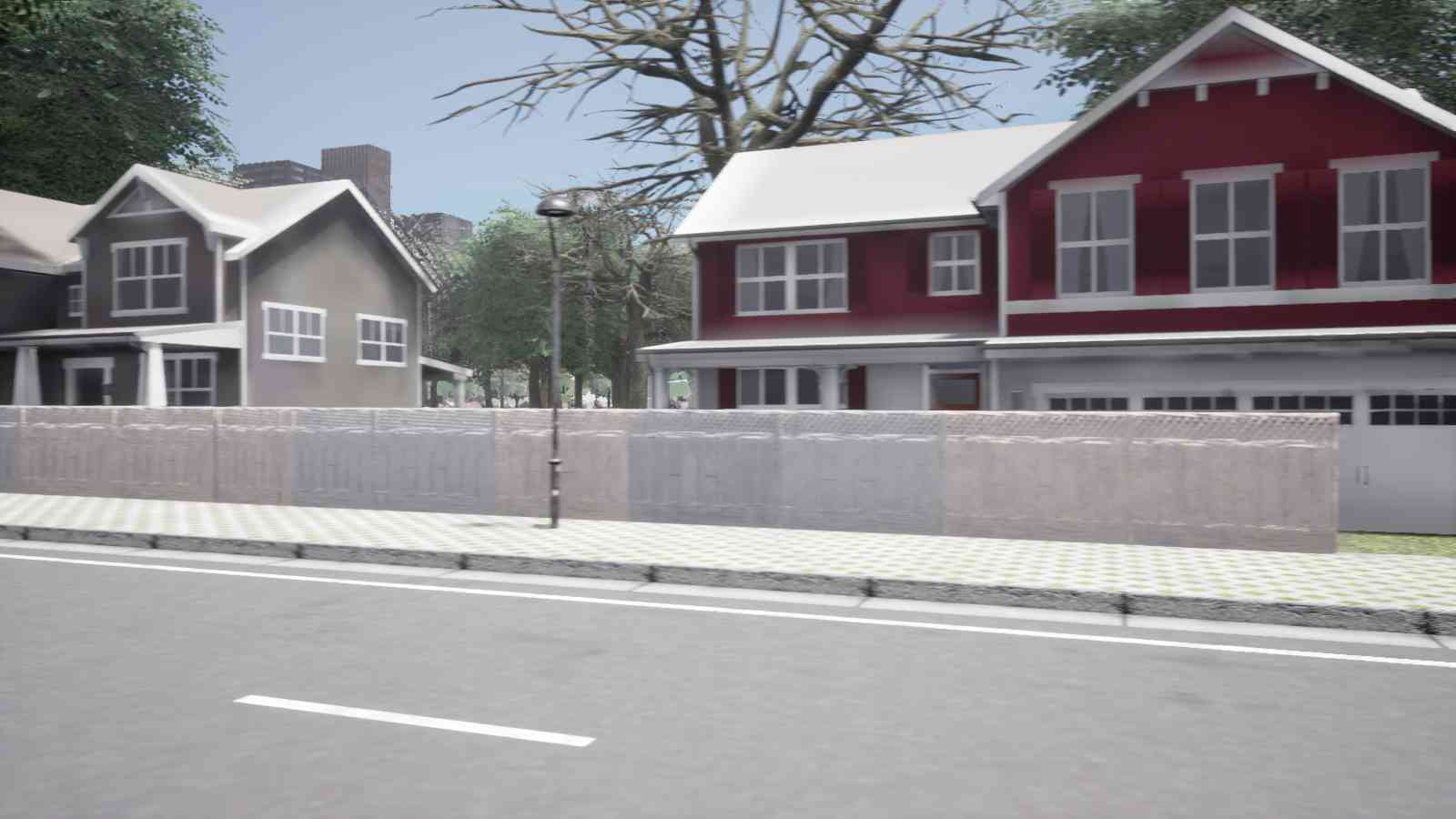} 
  \includegraphics[width=2.5cm]{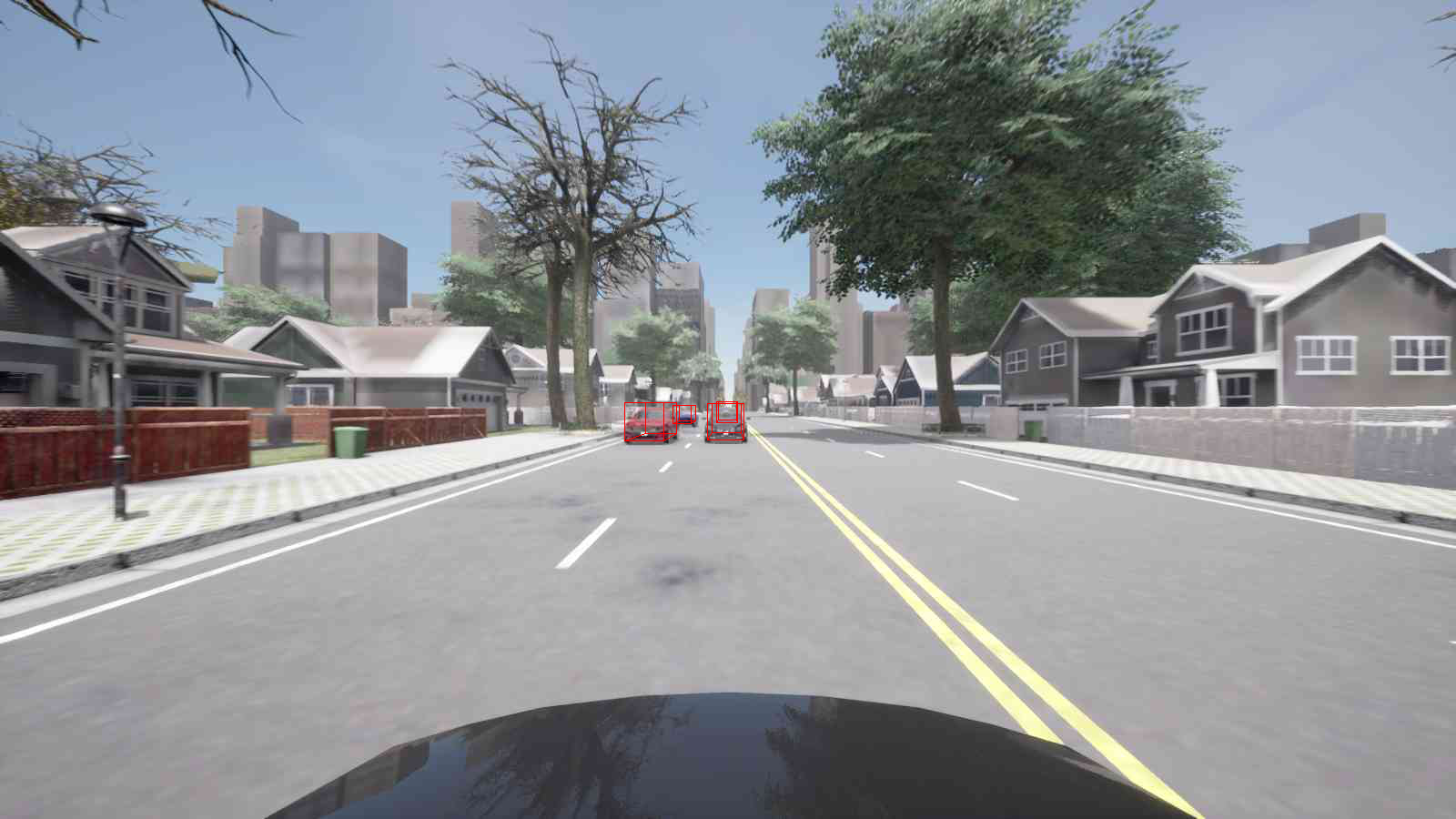} 
  \includegraphics[width=2.5cm]{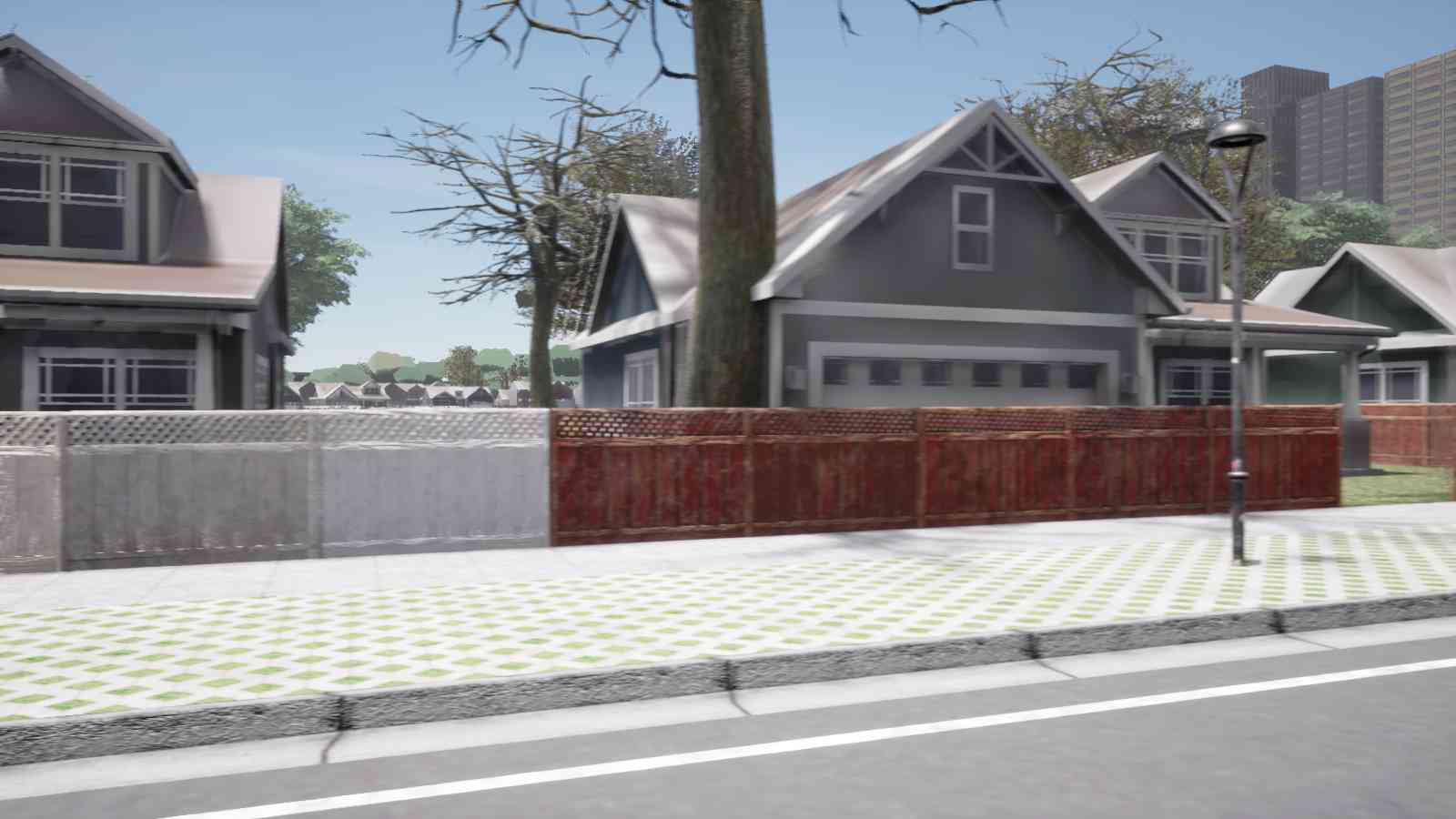} 
  \includegraphics[width=2.5cm]{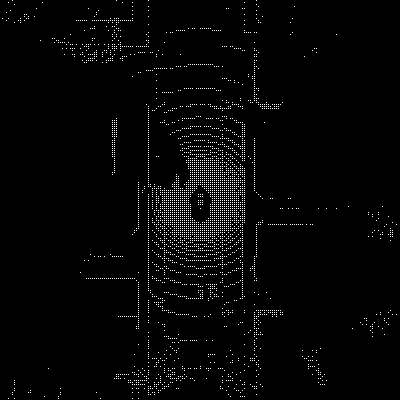} 
  \includegraphics[width=2.5cm]{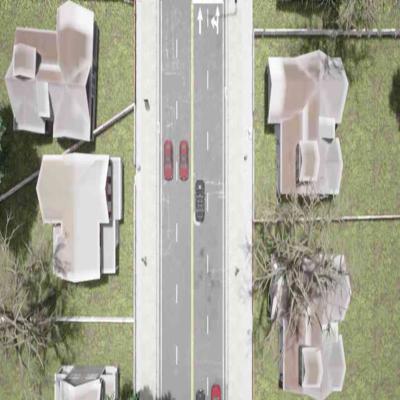}
  \includegraphics[width=2cm]{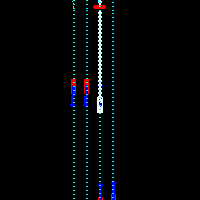} 
  \caption{These images include views from 6 cameras, point cloud from LiDAR sensors, and satellite-like BEV RGB image. We visualize the ground-truth 3D bounding boxes for vehicles, and 2D bounding boxes for traffic lights on camera-view. Additionally, ground-truths are visualized on a BEV grid, including centerlines, lane guidance, and orientation for vehicles.}
\end{figure}

In summary, our main contributions are as follows:
\begin{itemize}

    \item We introduce \textbf{TaCarla}, a new, large-scale dataset for the CARLA Leaderboard 2.0 challenge that addresses the gaps of previous works by providing a significantly \textbf{larger number of complex, multi-lane scenarios} requiring challenging lane-change maneuvers, as shown in Table~\ref{tab:scenario_comparison}.

    \item We \textbf{combine the strengths} of prior datasets by simultaneously using the robust \textbf{PDM expert policy} (addressing the oscillation issues found in Bench2Drive) and adopting the comprehensive \textbf{NuScenes sensor configuration} (providing the 360-degree coverage lacking in PDM-Lite).
    


    \item We provide a \textbf{holistic dataset} with rich annotations and, critically, a \textbf{comprehensive benchmark suite}. We establish robust SOTA baselines for a wide spectrum of tasks, covering both \textbf{perception components} (e.g., 3D object detection, centerline detection) and \textbf{end-to-end planning} models.


    \item We provide rule-based \textbf{text annotations} for each scene to support modern, LLM-based research. Furthermore, we introduce a normalized \textbf{rarity score} to quantify scenario uniqueness and help identify long-tail events in the dataset.
\end{itemize}
\section{Related work}
In alignment with the findings presented in \cite{kaplan2020scalinglawsneurallanguage}, larger models, when trained with increased data and computational resources, consistently surpass the performance of their smaller counterparts. This phenomenon underlies the widespread adoption and success of large language models. Inspired by these insights, we aim to achieve an order of magnitude increase in dataset size for each covered task compared to existing datasets in the literature. In this section, we analyize the sizes of the widely adopted datasets in autonomus driving and evaluate the range of tasks they address.


\paragraph{Dynamic object detection.} 

Dynamic object detection remains one of the foremost challenges in autonomous driving. The KITTI Object Detection Dataset~\cite{KITTI}, introduced in 2017, was specifically developed for 3D object detection and comprises 7,481 training images and 7,518 test images, encompassing a total of 80,256 labeled objects. The Waymo Open Dataset for perception tasks~\cite{waymo_open}, last updated in 2024, includes 390,000 frames for 3D object detection. The nuScenes dataset~\cite{caesar2020nuscenes}, released in 2021, features 1.4 million 3D bounding boxes and is widely regarded as one of the most popular datasets for dynamic object detection. Argoverse 2~\cite{wilson2023argoverse}, released in 2023, contains 150,000 samples. The nuPlan dataset~\cite{nuplan}, released in 2023 originally for planning, includes approximately 120 hours of raw sensor data with autolabeled ground-truths. 


\paragraph{2D Lane Detection.}
The 2D Lane Detection task aims to detect lane dividers from Perspective View (PV) and project them to a Bird's Eye View (BEV) under the flat world assumption. The TuSimple dataset~\cite{yoo2020endtoendlanemarkerdetection}, launched in 2017, contains 3,626 images in the training set, 358 in the validation set, and 2,782 in the test set. The CULane dataset \cite{pan2018SCNN}, released in 2018, comprises 133,235 frames, meticulously divided into 88,880 for the training set, 34,680 for the test set, and 9,675 for the validation set. However, both CULane and TuSimple datasets contain only four to five lane dividers, which limits the completeness of lane divider detection. The BDD100K dataset~\cite{yu2020bdd100kdiversedrivingdataset}, released in 2020, comprises 100,000 labeled images. 
The CurveLanes dataset~\cite{CurveLane-NAS}, published in 2020, includes 150,000 frames, with 100,000 allocated for training, 20,000 for validation, and 30,000 for testing. The Curvelanes dataset is important from the perspective that it includes high curvature lane dividers and includes up to 9 instances. Published in 2021, the VIL100 dataset~\cite{zhang2021vil100newdatasetbaseline} includes 10,000 frames. The CARLANE dataset~\cite{NEURIPS2022_19a26064}, released in 2022, is another dataset collected using Carla Simulation, focusing primarily on lane detection, and includes 84,000 images.

\paragraph{3D Lane Detection.}
The 3D Lane Detection task aims to detect the 3D Lane coordinates of lane dividers directly. These datasets address the limitation of flat world assumption in 2D lane detection datasets. Introduced in 2019, the LLamas dataset~\cite{llamas2019} encompasses a total of 100,000 annotated images. Published in 2020, the Gen-LaneNet dataset~\cite{Guo_2020} is a synthetic dataset and contains 6,000 samples. The KLane dataset~\cite{paek2022klane}, released in 2021, includes 15,000 frames of real images. The Once-3DLanes dataset~\cite{yan2022once}, released in 2022, includes 211,000 simulated samples. The OpenLane-V1 dataset~\cite{chen2022persformer}, launched in 2022, comprises 200,000 frames, offering a substantial resource for lane detection research. The LanEvil dataset~\cite{zhang2024lanevilbenchmarkingrobustnesslane}, published in 2024, was generated using the CARLA simulator and contains 90,292 images.

\paragraph{Centerline and Topology Dataset.}
The OpenLane-V2 dataset~\cite{wang2024openlane}, launched in 2024, comprises two distinct subsets, offering a comprehensive resource for lane detection research. Subset-A, derived from Argoverse V2, includes 22,477 training samples, 4,806 validation samples, and 4,816 test samples, with a resolution of 2048x1550. Subset-B, derived from NuScenes, contains 27,968 training samples, 6,019 validation samples, and 6,008 test samples, with a resolution of 1600x900.


\paragraph{Traffic Light Detection.}
Traffic light detection involves identifying 2D bounding boxes around traffic lights and determining their classes. This specialized task requires more detailed information and larger datasets than generic object detection~\cite{COCO, PascalVOC, LVIS, OpenImages}. The LISA Traffic Light Dataset~\cite{LisaTL} includes 46,418 frames and 112,971 annotated traffic lights with 7 classes. The BOSCH Small Traffic Light Dataset~\cite{BoschTL} has 13,427 images and approximately 24,000 annotated traffic lights. BDD100K~\cite{BDD100K} provides 100,000 driving videos and 14,606 annotated traffic lights, supporting various vision tasks for autonomous driving. Unlike these datasets, our contribution aligns traffic light data with other sensory inputs (LIDAR, Ego Status) and driving waypoint information.

\paragraph{Planning.} The aim of the planning task is to generate waypoints for the ego-vehicle to navigate to its destination while addressing complex challenges. Several datasets have been developed to address this problem.
The NuPlan dataset~\cite{nuplan}, comprising 1,200 hours of driving data, is one of the pioneering benchmarks in this domain. It includes dynamic objects, traffic lights on the HD map, and lane information from the HD map. However, a significant limitation of the NuPlan dataset is its lack of diversity in planning tasks, as a large portion of the data involves the ego vehicle merely moving forward \cite{hallgarten2024vehiclemotionplanninggeneralize, zhang2025cafeadcrossscenarioadaptivefeature, chitta2024sledgesynthesizingdrivingenvironments}. 
The Bench2Drive dataset~\cite{bench2drive}, released for the Leaderboard 2.0 challenge in 2024, contains 2 million frames collected at 10 Hz using a reinforcement learning (RL)-based expert that leverages ground truth information from the CARLA simulation. A drawback of this dataset is the potential for oscillation issues in the ego vehicle, which can affect its behavior in scenarios—a common problem in RL agents~\cite{chen2021addressingactionoscillationslearning, mysore2021regularizingactionpoliciessmooth}. In our dataset, the PDM-Lite rule-based expert was used to collect the data.
The PDM-Lite dataset, published in 2024, was created using the PDM-Lite rule-based expert. It contains 581,662 samples, collected at a frequency of 4 Hz. However, this dataset is limited by its sensor configuration. In contrast, our dataset utilizes one of the popular sensor configurations from the NuScenes dataset \cite{caesar2020nuscenes}, allowing for a seamless switch between the two.

\section{TaCarla}

\begin{figure}[h]
  \centering
  \begin{subfigure}[t]{0.35\textwidth}
    \includegraphics[width=\linewidth]{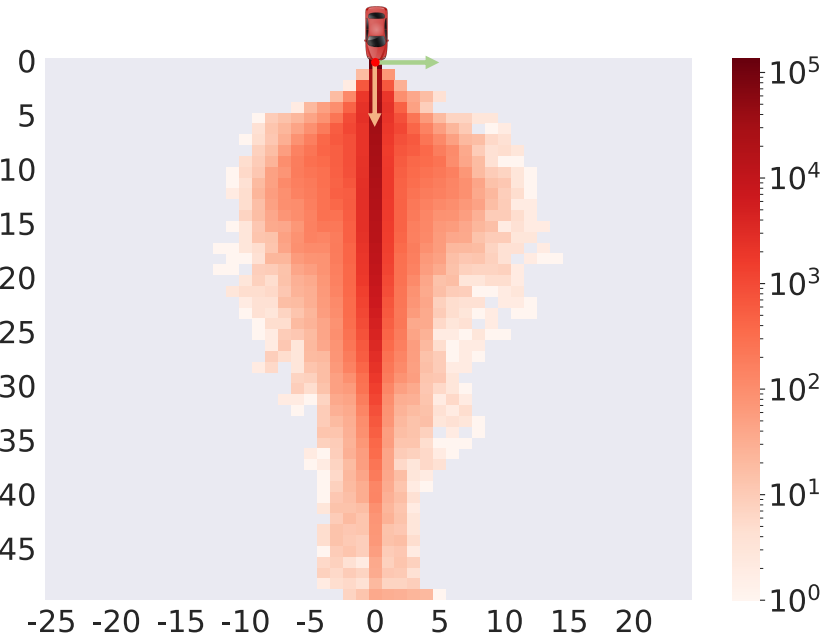}
    \caption{nuScenes \cite{caesar2020nuscenes} heatmap drawn by ~\cite{is_ego_all_you_need}}
    \label{fig:ego-dist-nuscenes}
  \end{subfigure}
  \hfill
  \begin{subfigure}[t]{0.31\textwidth}
    \includegraphics[width=\linewidth]{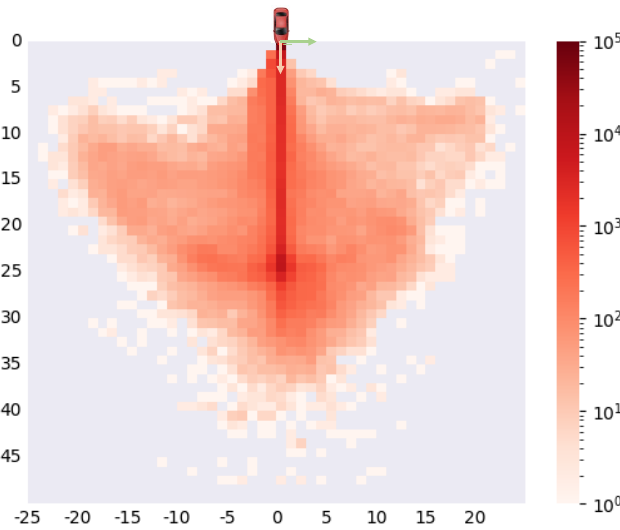}
    \caption{Bench2Drive \cite{bench2drive} heatmap.}
    \label{fig:ego-dist-b2d}
  \end{subfigure}
  \hfill
  \begin{subfigure}[t]{0.31\textwidth}
    \includegraphics[width=\linewidth]{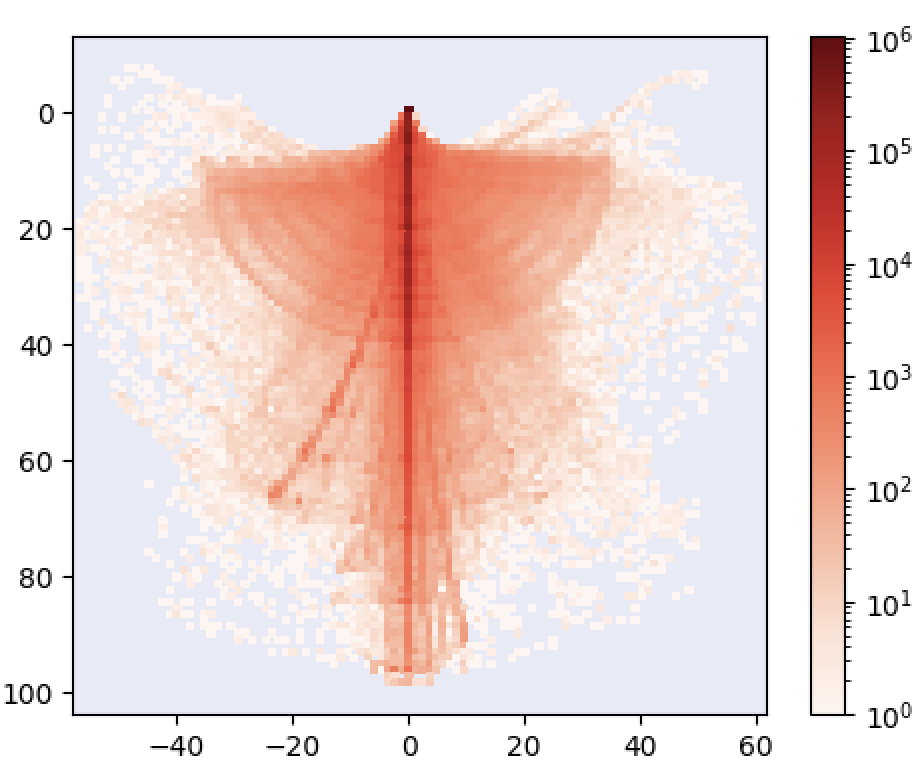}
    \caption{Tacarla heatmap.}
    \label{fig:ego-dist-tacarla}
  \end{subfigure}
  
  \caption{Distribution of the ego vehicle’s ground-truth location.}
  \label{fig:ego-future-distribution}
\end{figure}

Samples of the TaCarla dataset were meticulously collected based on training and validation routes provided by the Leaderboard 2.0 challenge in the CARLA 0.9.15 simulation. The Leaderboard 2.0 encompasses 36 distinct scenarios, including lane changing at the appropriate time, interacting with pedestrians, prioritizing emergency vehicles, managing traffic lights, and responding to unexpected stops of nearby vehicles, among others, to emulate real-life traffic situations. Detailed descriptions of these scenarios are provided in the Appendices. To demonstrate the diversity of trajectories in our dataset, we compare it with nuScenes~\cite{caesar2020nuscenes} and Bench2Drive~\cite{bench2drive} in Figure~\ref{fig:ego-future-distribution}, in a similar fashion to the comparison provided in Bench2Drive~\cite{bench2drive}. This comparison illustrates that our data collection expert can execute more diverse movements over extended distances. Our heatmap, depicting trajectories over a 4-second horizon, indicates that our vehicle can reach speeds of up to 72 km/h, thereby showcasing the diversity in velocity as well. While the trajectory heatmap in Figure \ref{fig:ego-future-distribution} demonstrates greater spatial reach and velocity diversity compared to nuScenes and Bench2Drive, we note that the PDM rule-based expert introduces its own distributional biases. Specifically, the policy tends to operate within a moderate speed regime, leading to visible density bands at certain velocity levels rather than a smooth continuum. Additionally, the route structure of the CARLA Leaderboard 2.0 XML files results in a higher frequency of right-turn maneuvers compared to left-turns in the collected data. 



Given the multitude of scenarios in the XML files, we meticulously separated them into distinct routes based on the trigger points where each scenario is executed. This separation was performed between two consecutive scenarios on the original routes. The dataset comprises over 2,850,000 frames recorded at 10 Hz under various weather conditions, utilizing the NuScenes sensor configuration, which is one of the most widely used setups for perception tasks. Therefore, akin to the NuScenes dataset, our dataset incorporates 6 RGB cameras, 5 radars, and 1 LiDAR sensor, all positioned identically to those in NuScenes. In addition to these sensors, we also collect bird’s-eye view RGB images (reminiscent of satellite imagery), depth camera data, instance segmentation images, and semantic segmentation images, as these modalities are extensively utilized in perception research. Furthermore, depth, instance segmentation, and semantic segmentation images are captured using the same camera configuration, aligning with common practices in the literature for perception tasks.

The dataset encompasses seven distinct object types, categorized based on their significance in planning scenarios. These categories include \textit{walker, car, police, ambulance, firetruck, crossbike,} and \textit{construction}. Specifically, the dataset comprises 67,985 walkers, 7,939,572 cars, 115,022 police vehicles, 9,442 ambulances, 3,972 firetrucks, 119,165 crossbikes, and 427,622 construction objects. Additionally, there are 238,780 traffic light samples in Town12 and 187,987 in Town13.

The TaCarla dataset includes environmental information such as initial weather conditions stored in the \texttt{simulation\_results} in the corresponding parquet file. 
The key weather parameters are \textit{cloudiness}, \textit{fog\_density}, \textit{precipitation}, \textit{precipitation\_deposits}, and \textit{wetness}. 
Each parameter ranges from 0 to 100.

In Table\ref{tab:weather-distribution}, to improve interpretability, these numerical values are grouped into categories: 
\textit{very\_low\_effect} (0--20), 
\textit{low\_effect} (20--40), 
\textit{medium\_effect} (40--80), and 
\textit{heavy\_effect} (80--100).

\begin{table}[htbp]
\centering
\caption{Weather Condition Distributions in the TaCarla Dataset}
\scriptsize 
\setlength{\tabcolsep}{4pt} 
\renewcommand{\arraystretch}{0.9} 
\resizebox{1.0\columnwidth}{!}{ 
\begin{tabular}{lcccc}
\hline
\textbf{Condition} & \textbf{Clean (\%)} & \textbf{Low (\%)} & \textbf{Medium (\%)} & \textbf{Heavy (\%)} \\
\hline
Cloudiness & 35.49 & 7.65 & 34.22 & 22.62 \\
Fog Density & 68.45 & 3.31 & 23.82 & 4.40 \\
Precipitation & 51.77 & 3.76 & 33.99 & 10.46 \\
Precipitation Deposits & 33.65 & 3.37 & 49.16 & 13.80 \\
Wetness & 79.87 & 0.41 & 10.79 & 8.91 \\
\hline
\end{tabular}}
\label{tab:weather-distribution}
\end{table}


In some scenarios, such as Accident, AccidentTwoWays, ConstructionObstacle, ConstructionObstacleTwoWays, HazardAtSideLane, HazardAtSideLaneTwoWays, ParkedObstacle, ParkingCrossingPedestrian, ParkingExit, and YieldToEmergencyVehicle, the ego vehicle needs to change lanes to accomplish the scenario. Therefore, it must observe not only the vehicle in front of it but also other vehicles in adjacent lanes. Consequently, these scenarios are more complex than the rest. The comparison of scenarios in TaCarla, Bench2Drive, and PDM-Lite is presented in Table~\ref{tab:scenario_comparison}. It can be observed that the number of such tasks in TaCarla significantly exceeds those in the other datasets. The details of these scenarios can be found in the Appendix.

\begin{table}[htbp]
\centering
\caption{Number of scenarios in TaCarla compared to Bench2Drive and PDM-Lite}
\resizebox{\columnwidth}{!}{%
\begin{tabular}{lccc}
\hline
\textbf{Scenario} & \textbf{TaCarla} & \textbf{Bench2Drive} & \textbf{PDM-Lite} \\
\hline
Accident & \textbf{353} & 28 & 88 \\
AccidentTwoWays & \textbf{410} & 201 & 97 \\
BlockedIntersection & 43 & \textbf{582} & 96 \\
ConstructionObstacle & \textbf{306} & 27 & 92 \\
ConstructionObstacleTwoWays & \textbf{310} & 138 & 94 \\
ControlLoss & 255 & 18 & \textbf{266} \\
CrossingBicycleFlow & \textbf{153} & 105 & 50 \\
DynamicObjectCrossing & \textbf{452} & 26 & 266 \\
EnterActorFlow & 71 & 19 & \textbf{112} \\
HardBreakRoute & \textbf{281} & 35 & 96 \\
HazardAtSideLane & 123 & \textbf{187} & 78 \\
HazardAtSideLaneTwoWays & \textbf{358} & 27 & 97 \\
HighwayCutIn & \textbf{144} & 39 & 83 \\
HighwayExit & 28 & 29 & \textbf{88} \\
InterurbanActorFlow & 10 & 13 & \textbf{92} \\
InterurbanAdvancedActorFlow & 79 & 42 & \textbf{84} \\
MergerIntoSlowTraffic & 118 & 71 & \textbf{173} \\
NonSignalizedJunctionLeftTurn & \textbf{308} & 116 & 99 \\
NonSignalizedJunctionRightTurn & \textbf{391} & 156 & 91 \\
OppositeVehicleRunningRedLight & \textbf{385} & 85 & 261 \\
OppositeVehicleTakingPriority & 231 & \textbf{614} & 99 \\
ParkedObstacle & \textbf{261} & 28 & 93 \\
ParkedObstacleTwoWays & \textbf{416} & 23 & 90 \\
ParkingCrossingPedestrian & \textbf{338} & 20 & 90 \\
ParkingCutIn & \textbf{259} & 38 & 96 \\
ParkingExit & \textbf{132} & 27 & 98 \\
PedestrianCrossing & 188 & \textbf{215} & 99 \\
PriorityAtJunction & \textbf{492} & 116 & 94 \\
SignalizedJunctionLeftTurn & 185 & 21 & \textbf{327} \\
SignalizedJunctionRightTurn & 213 & 15 & \textbf{237} \\
StaticCutIn & \textbf{230} & 24 & 100 \\
VehicleTurningRoute & 124 & 235 & \textbf{584} \\
VehicleTurningRoutePedestrian & 87 & 18 & \textbf{98} \\
YieldToEmergencyVehicle & \textbf{268} & 19 & 84 \\
\hline
\end{tabular}%
}
\label{tab:scenario_comparison}
\end{table}

Since large language models (LLMs) have become very prevalent recently, many researchers have started incorporating them into autonomous driving tasks to enhance reasoning abilities of AI models. Therefore, in this study, we also generate textual descriptions that provide information about the current scene of the ego vehicle. These text annotations were extracted using a rule-based approach that leverages the 3D labels of lanes, objects, and lane guidance in our dataset.

In addition, the CARLA Leaderboard 2.0 challenge includes many edge-case scenarios. However, as shown in the distributions in Figure \ref{fig:ego-future-distribution}, the ground truths for simply following the route and going straight are overrepresented in all three datasets. As one might expect, this can lead to an imbalance in the datasets, as the ego vehicles may not effectively learn other behaviors. To quantify this imbalance and identify rare events, we propose a \textbf{rarity score} applicable to any dataset containing text annotations. This score adapts the Inverse Document Frequency (IDF)~\cite{idf} to measure the uniqueness of a given text annotation $W_t$ (representing the current scenario) relative to the entire corpus $N$. The score is calculated as:



 \[
 \mathrm{Rarity}(W_t) = \frac{1}{|W_t|} \sum_{w \in W_t} \log\!\left( \frac{1+l_N}{1 + \sum_{n \in N} \mathbf{1}_{\{\, w \in n \,\}}} \right)
 \]

\[
\mathrm{FinalRarity}(W_t) = 
\frac{\mathrm{Rarity}(W_t) - \min\!\left(\mathrm{Rarity}\right)}
     {\max\!\left(\mathrm{Rarity}\right) - \min\!\left(\mathrm{Rarity}\right)}
\]

\noindent
where $\mathrm{Rarity}(W_t)$ represents the unnormalized rarity value of the current text $W_t$, and $\min(\mathrm{Rarity})$ and $\max(\mathrm{Rarity})$ denote the minimum and maximum rarity scores observed across the entire dataset. This normalization maps all rarity values into the range $[0,1]$, where higher values indicate rarer or more distinctive textual descriptions.

In this formula, $l_N$ is the total number of sentences (corpus length), $W_t$ is the current sentence being scored, $w$ is a word in $W_t$, $n$ is any sentence in the corpus $N$, and $\mathbf{1}_{\{\cdot\}}$ is the indicator function.

\begin{figure}[h]
  \centering
  \begin{subfigure}[t]{0.37\textwidth}
    \includegraphics[width=\linewidth]{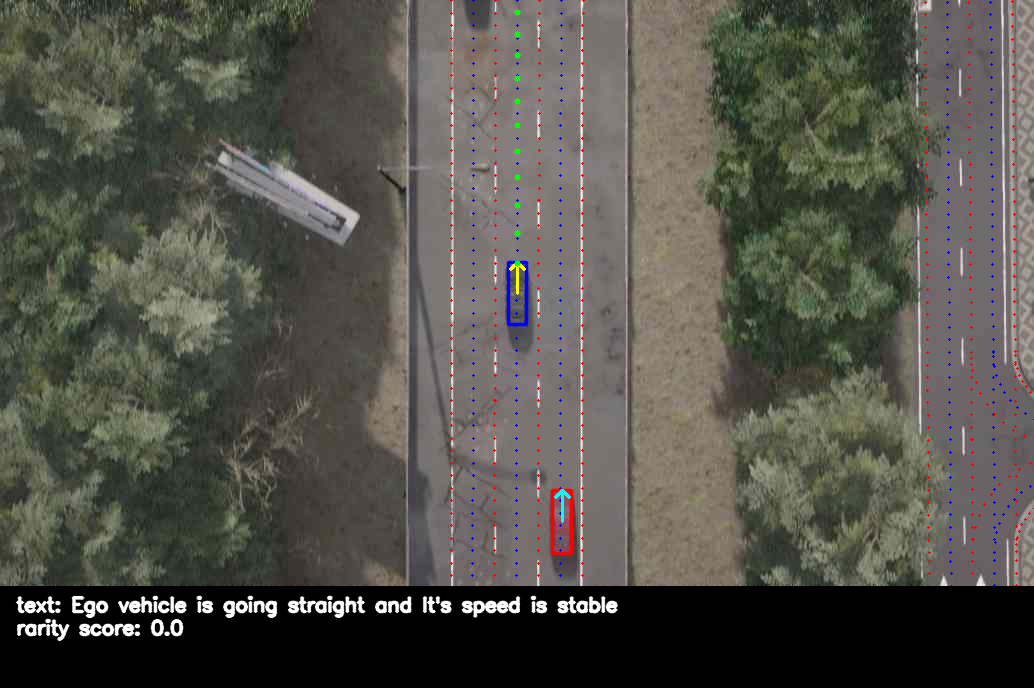}
    \caption{The ego vehicle is just following the route.}
    \label{fig:ego-dist-nuscenes}
  \end{subfigure}
  \hfill
  \begin{subfigure}[t]{0.37\textwidth}
    \includegraphics[width=\linewidth]{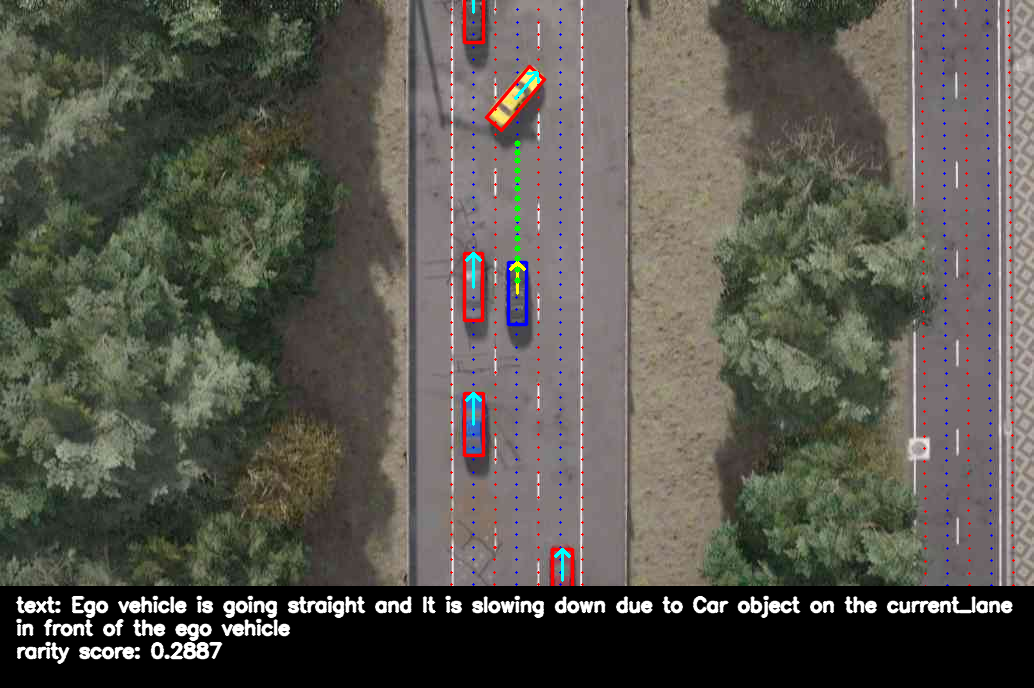}
    \caption{The ego vehicle is slowing down because of another car.}
    \label{fig:ego-dist-b2d}
  \end{subfigure}
  \hfill
  \begin{subfigure}[t]{0.37\textwidth}
    \includegraphics[width=\linewidth]{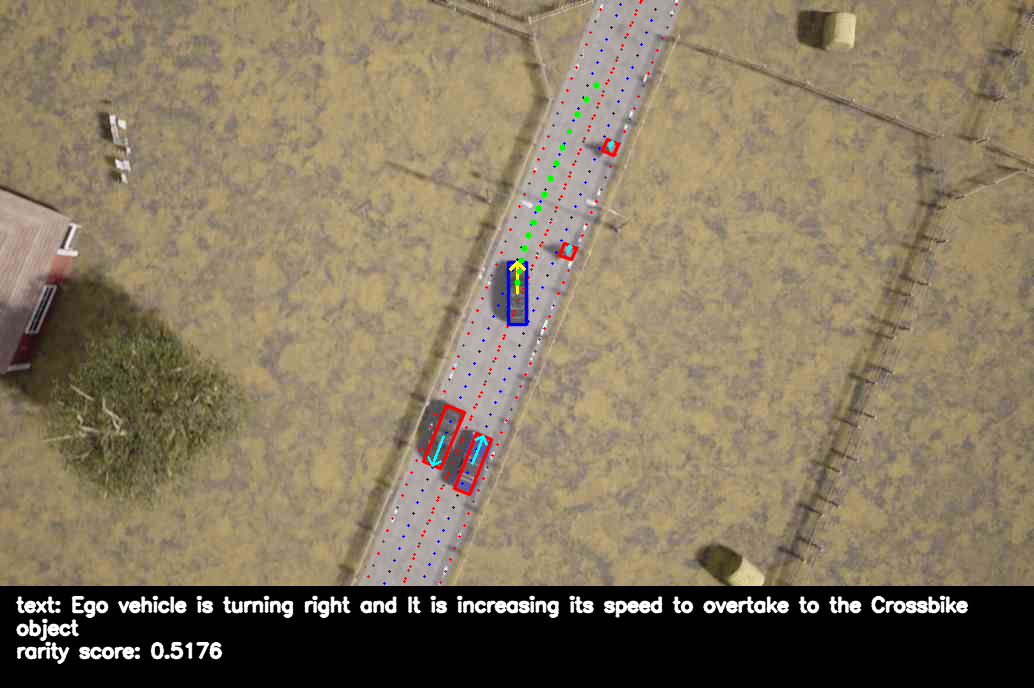}
    \caption{The ego vehicle is overtaking the bicycles.}
    \label{fig:ego-dist-tacarla}
  \end{subfigure}
  
  \caption{Outputs of the TaCarla Label Viewer}
  \label{fig:text}
\end{figure}

Empirically, this score effectively distinguishes common from rare events. As illustrated in Figure~\ref{fig:text}, common route-following scenarios register a low score (approx. $0.0$), whereas the score increases significantly for more complex or unusual situations.


\section{Experiments}
\label{sec:experiments}
\subsection{3D Object Detection}
We employed a non-transformer-based architecture for multi-view bird’s eye view (BEV)-based 3D object detection. In this architecture, the multi-view camera images are initially processed by a convolutional image encoder, specifically RegNetY-800MF \cite{RegNet}, with a feature pyramid network based on BiFPN \cite{tan2020efficientdet}. We project the feature levels at /8, /16, and /32 resolutions into the BEV representation using Lift-Splat projection \cite{LSS}. Features from the previous two frames are warped to the current frame using egomotion and then concatenated along the channel dimension, similar to BevDet4D \cite{huang2022bevdet4d}. Gradients produced by the previous frames are not used to update the image encoder. The resulting spatio-temporal BEV features are processed by a ResNet-based \cite{resnet34} BEV backbone. These features are then shared among task-specific heads. We utilize RQR3D \cite{rqr3d} for BEV-based 3D object detection. RQR3D reparametrizes the regression targets for the 3D bounding boxes and implements this reparameterized regression task on an anchor-free single-stage object detector, introducing an objectness head to address class imbalance problems of single-stage object detectors. RQR3D outperforms widely-adopted CenterPoint--based approaches \cite{yin2021center}, yielding lower translation and orientation errors, which are crucial for safe autonomous driving. When using LiDAR, we simply map the point cloud onto the BEV grid and concatenate it with the projected image feature before temporal processing.

To align our dataset with the nuScenes \cite{caesar2020nuscenes} benchmark, which provides annotated keyframes at 2 Hz, we downsample our original 10 Hz data to 2 Hz. This conversion ensures consistency in temporal resolution, facilitating fair comparisons and compatibility with existing evaluation protocols. 
Additionally, we select the scenarios whose name contains keywords such as \textit{accident, construction, dynamic, pedestrian, hazard, emergency,} and \textit{opposite} in order to achieve a more balanced class distribution. 
 
We utilize two different sensor configurations: camera-only and camera-LiDAR. Evaluation metrics are adopted from nuScenes \cite{caesar2020nuscenes}, including mean Average Precision (mAP), Average Translation Error (ATE), Average Scale Error (ASE), Average Orientation Error (AOE) and Average Velocity Error (AVE), excluding Average Attribute Error (AAE) as it is not applicable for TaCarla. These metrics provide a comprehensive assessment of detection performance across various object classes. The inclusion of LiDAR modality enhances depth estimation accuracy, leading to improved localization and orientation predictions, as reflected in lower ATE and AOE values. Conversely, the camera-only configuration exhibits higher errors due to the inherent challenges in depth estimation. The detailed class-wise performance metrics are presented in Tables \ref{tab:3d_results_lss} and \ref{tab:3d_results_lss_lidar}, illustrating the comparative effectiveness of both training approaches.

\begin{table}[t]
  \caption{Camera-only 3D object detection performance.}
  \label{tab:3d_results_lss}
  \centering
  \resizebox{\columnwidth}{!}{
  \begin{tabular}{@{}lccccc@{}}
    \toprule
                      & \textbf{AP} & \textbf{ATE} & \textbf{ASE} & \textbf{AOE} & \textbf{AVE} \\ \midrule
    \textbf{Car}      & 0.459       & 0.444        & 0.147        & 0.012        & 0.559        \\
    \textbf{Crossbike}& 0.324       & 0.242        & 0.094        & 0.057        & 0.165        \\
    \textbf{Walker}   & 0.426       & 0.456        & 0.885        & 1.333        & 0.292        \\
    \textbf{Police}   & 0.381       & 0.249        & 0.056        & 0.011        & 0.048        \\
    \textbf{Construction} & 0.419   & 0.665        & 0.812        & 1.125        & 0.065        \\
    \textbf{Ambulance}& 0.098       & 0.440        & 0.132        & 0.065        & 0.525        \\
    \textbf{Firetruck}& 0.140       & 0.487        & 0.155        & 0.004        & 0.618        \\ \midrule
                          \multicolumn{1}{@{}l}{\textbf{Mean}} 
      & \textbf{0.32} 
      & \textbf{0.43} 
      & \textbf{0.33} 
      & \textbf{0.37} 
      & \textbf{0.32} \\
    \bottomrule
  \end{tabular}
  }
\end{table}

\begin{table}[t]
  \caption{Camera-LiDAR 3D object detection performance.}
  \label{tab:3d_results_lss_lidar}
  \centering
  \resizebox{\columnwidth}{!}{
  \begin{tabular}{@{}lccccc@{}}
    \toprule
                      & \textbf{AP} & \textbf{ATE} & \textbf{ASE} & \textbf{AOE} & \textbf{AVE} \\ \midrule
    \textbf{Car}      & 0.716       & 0.173        & 0.125        & 0.022        & 0.399        \\
    \textbf{Crossbike}& 0.556       & 0.113        & 0.086        & 0.076        & 0.119        \\
    \textbf{Walker}   & 0.527       & 0.152        & 0.885        & 1.304        & 0.277        \\
    \textbf{Police}   & 0.486       & 0.082        & 0.052        & 0.018        & 0.038        \\
    \textbf{Construction} & 0.657   & 0.253        & 0.821        & 1.125        & 0.098        \\
    \textbf{Ambulance}& 0.428       & 0.254        & 0.100        & 0.074        & 0.283        \\
    \textbf{Firetruck}& 0.452       & 0.270        & 0.108        & 0.002        & 0.344        \\ \midrule
                      \multicolumn{1}{@{}l}{\textbf{Mean}} 
      & \textbf{0.55} 
      & \textbf{0.19} 
      & \textbf{0.31} 
      & \textbf{0.37} 
      & \textbf{0.22} \\
    \bottomrule
  \end{tabular}
  }
\end{table}


\subsection{Lane Detection}
\begin{table}[t]
  \caption{Centerline and Lane Divider Detection Results of TopoBDA architecture for TaCarla Dataset.}
  \label{tab: topobda_tacarla}
  \centering
  \begin{tabular}{lccc}
    \toprule
    \textbf{Detection Task} & \textbf{\( \text{AP}_f \)} & \textbf{\( \text{AP}_c \)} & \textbf{F1\(_{1.5}\)} \\
    \midrule
    Centerline Detection & 39.6 & 41.7 & 67.3 \\
    Lane Divider Detection & N/A & 32.1 & 64.3 \\
    \bottomrule
  \end{tabular}
\end{table}

Lane detection consists of two sub-tasks which are lane divider detection and centerline detection. For the lane divider and centerline detection tasks, Chamfer Distance-based Average Precision (\( \text{AP}_c \)) \cite{li2022hdmapnet, liao2024maptrv2} and Frechet Distance-based Average Precision (\( \text{AP}_f \)) \cite{wang2024openlane, li2023graph, wu2023topomlp} metrics were utilized. These metrics are commonly used in the literature to evaluate the geometric similarity between predicted and ground-truth polylines. For \( \text{AP}_f \), the thresholds are 1, 2, and 3 meters, and for \( \text{AP}_c \), the thresholds are 0.5, 1, and 1.5 meters. Additionally, the F1 metric is included, which is another widely used metric in the literature for assessing lane detection performance \cite{Guo_2020, chen2022persformer}. In the F1 metric calculation, if 75\% of the points are within the predetermined threshold, the instances are assumed to be true positives. In this study, this threshold is set to 1.5 meters, in accordance with the literature. For all evaluation metrics, 11 ground truth points are utilized.

For the training of both centerlines and lane dividers, the TopoBDA study \cite{kalfaoglu2024topobda} was utilized, which incorporates specialized attention structures and advanced polyline training practices derived from the TopoMaskV2 study \cite{kalfaoglu2024topomaskv2}. The Bezier Deformable Attention mechanism, a key innovation of TopoBDA, focuses attention around Bezier keypoints rather than a single central point. This approach enhances the efficiency of polyline learning by improving the detection and representation of elongated and thin polyline structures.

\begin{figure}[t]
  \centering
  \includegraphics[width=1.0\linewidth]{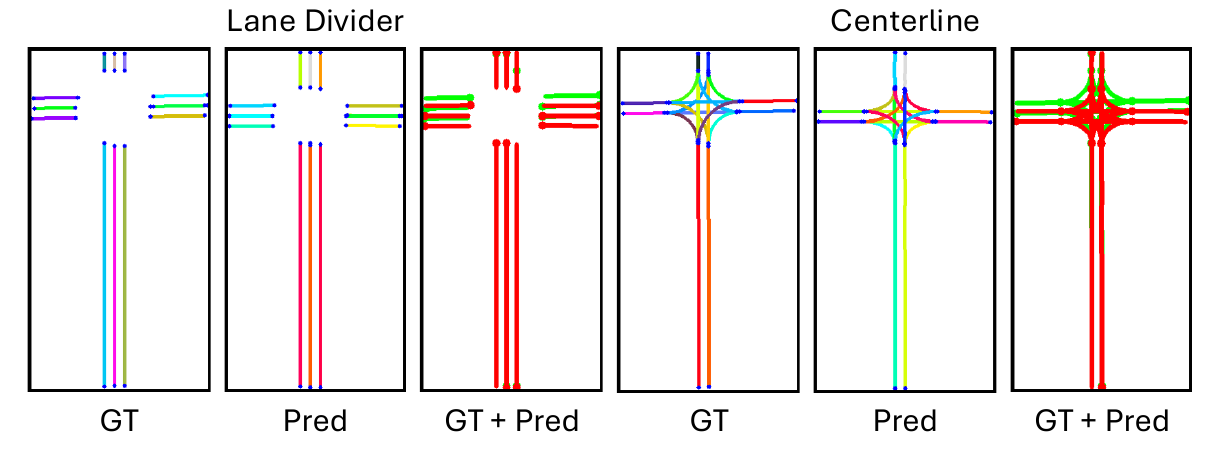}
  \caption{Bird's Eye View (BEV) results demonstrating the performance of TopoBDA on the TaCarla dataset. GT denotes the ground truth, and Pred denotes the predictions. GT + Pred shows the overlaid results of both, facilitating visual assessment of localization performance.}
  \label{fig: bev_samples}
\end{figure}

The results for the training of TopoBDA for lane divider and centerline detection are presented in Table \ref{tab: topobda_tacarla}. The experimental details for this training are consistent with those outlined in the TopoBDA study \cite{kalfaoglu2024topobda} except that the training duration is set to 6 epochs. The Frechet Distance-based Average Precision (\( \text{AP}_f \)) is marked as N/A because the Frechet distance emphasizes directional information, which is not relevant to the lane divider detection task. In the experiments, the TaCarla dataset is utilized in a 2Hz configuration, and the validation set is subsampled further with a factor of 5. 

Bird's Eye View (BEV) demonstrations are presented in Figure \ref{fig: bev_samples}, respectively. These figures include visualizations of both centerlines and lane dividers. In the figure, GT denotes the ground truth, while Pred represents the predictions made by TopoBDA. Additionally, Figure \ref{fig: bev_samples} includes an overlaid visualization (GT + Pred), which aids in understanding the localization performance. It is noteworthy that in junction regions, lane dividers are excluded, whereas centerlines are included, as illustrated in Figure \ref{fig: bev_samples}.

\subsection{Traffic Light Detection}
The dataset we propose consists of 238,780 and 187,987 images containing traffic light instances in the training and validation sets, respectively. Every individual image in the dataset consists of a traffic light instance with three distinct classes: \textit{red}, \textit{yellow}, \textit{green}. Every instance is labeled with its 2D bounding box and corresponding class. For the traffic light detection task, we employed off the shelf single-stage object detector FCOS \cite{fcos} with ResNet-50 backbone as a baseline. We use a 1x training schedule; we train for 12 epochs with a learning rate of $1e^{-3}$. We schedule the learning rate at 8th and 11th epochs with a rate of $0.1$. We report COCO style \cite{COCO} $\mathrm{AP}$ and $\mathrm{AP_{50}}$ in Table \ref{tab:TLR}.

\begin{table}[th]
  \caption{Traffic Light Detection task results in TaCarla.}
  \label{tab:TLR}
  \centering
  \begin{tabular}{lcc}
    \toprule
    \textbf{Model} & $\mathrm{AP}$ & $\mathrm{AP_{50}}$ \\
    \midrule
    FCOS \cite{fcos} & $59.5$ & $88.2$ \\
    \bottomrule
  \end{tabular}
\end{table}

\subsection{Planning}

\begin{figure}[h]
  \centering
  \includegraphics[width=8cm]{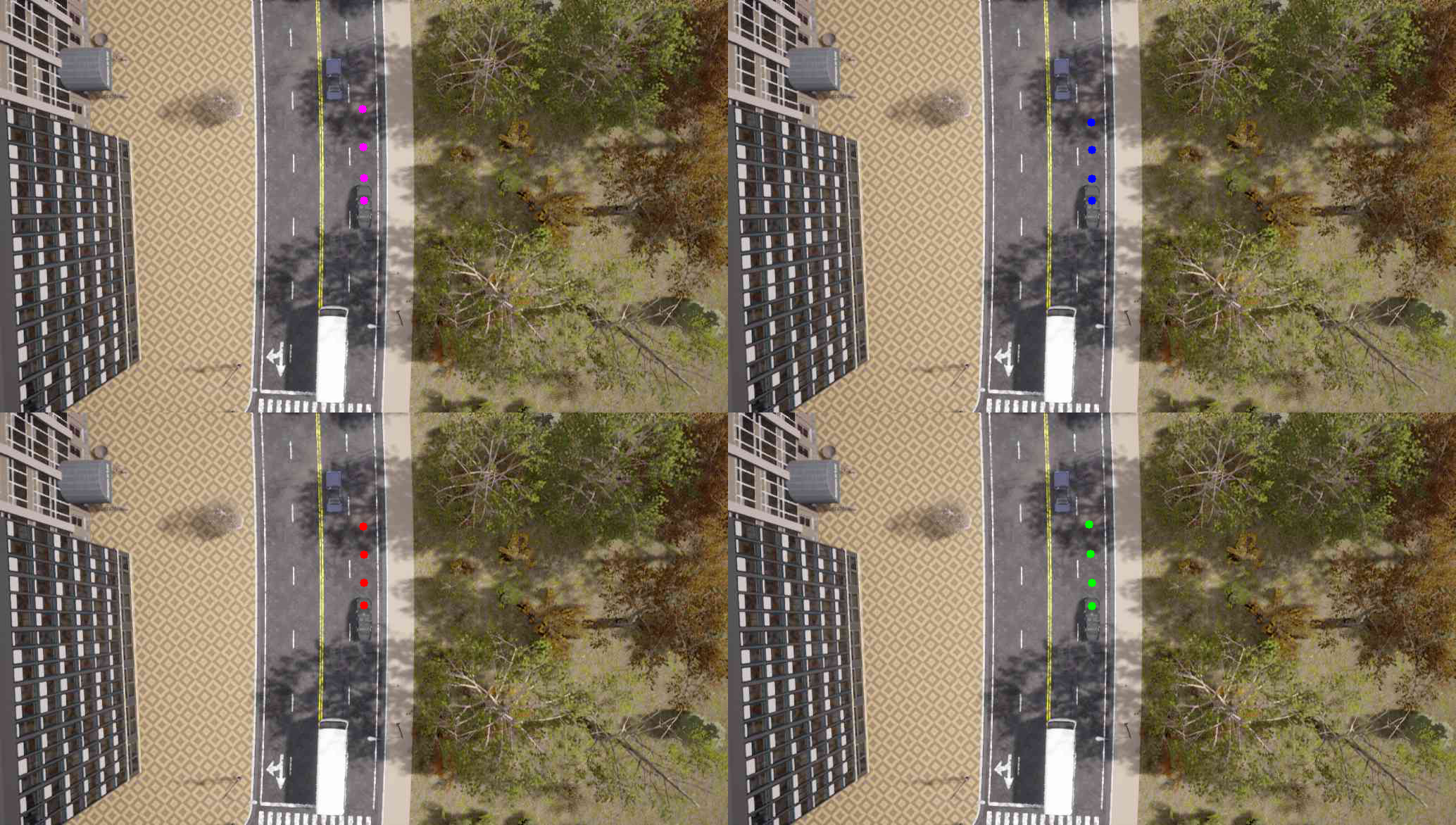} 
  \caption{Waypoints from ground truth(Top Left), PlanT(Top Right), Transfuser(Bottom Left), and DiffusionDrive (Bottom Right) models.}
\end{figure}

\begin{table}
  \caption{Open-loop results on the validation set of unseen Town13 with 1s, 2s, and 4s planning horizons (H).}
  \label{tab:openloop-results}
  {
  \begin{tabular}{l c | cccc}
    \toprule
    Model & $\mathcal{H}$ & ADE & FDE & AHE & FHE \\
    \midrule
    DiffusionDrive \cite{diffusiondrive} & 4 & 2.69 & 5.58 & 0.27 & 0.21 \\
                                         & 2 & 1.14 & 2.14 & 0.21 & 0.21 \\
                                         & 1 & 0.51 & 0.78 & 0.12 & 0.12 \\
    \midrule
    PlanT \cite{planT}                   & 4 & - & - & - & - \\
                                         & 2 & 1.03 & 1.71 & 0.36 & 0.34 \\
                                         & 1 & - & - & - & - \\
    \midrule
    Transfuser \cite{transfuser}         & 4 & 2.29 & 4.97 & 0.23 & 0.27 \\
                                         & 2 & 0.91 & 1.74 & 0.23 & 0.27 \\
                                         & 1 & 0.40 & 0.59 & 0.22 & 0.22 \\
    \bottomrule
  \end{tabular}
  }
\end{table}

For the planning task, we trained three baseline agents: Transfuser~\cite{transfuser}, DiffusionDrive~\cite{diffusiondrive}, and PlanT~\cite{planT}. 
We choose Transfuser and DiffusionDrive upon their great success in the Navsim~\cite{navsim} dataset. Both agents were trained with the same Resnet-34 backbone~\cite{resnet34}, which uses 3 forward-facing cameras and LiDAR. The cameras are cropped and concatenated as a single image of size $256x1024$ and LiDAR point clouds rasterized with a BEV size $256x256$. Both agents use the same ego status input, which consists of vectorel velocity, acceleration, and driving command. The driving command is calculated as it is in Navsim~\cite{navsim}. We used our annotated lane guidance waypoints by classifying the point 15m away from the ego position as left, right, or straight by checking the lateral distance with a threshold of 2m. We trained DiffusionDrive for 6 epochs and Transfuser for 3 epochs with a learning rate of $7.5 \times 10^{-5}$ on 8 NVIDIA A100 GPUs with a total batch size of 64. We filtered our training set scenarios with a driving score >70. During training, we sampled ground truth trajectories with 2Hz, which gives 8 waypoints for a 4 seconds horizon. The implementation architecture is the same with DiffusionDrive paper~\cite{diffusiondrive}, where we used 20 anchors clustered from our dataset. During evaluation, we used 2 denoising steps as the authors used in the Navsim challenge~\cite{navsim}. The other chosen planning model was the PlanT model \cite{planT} because it is one of the best planning models that utilizes ground truth information. We trained the PlanT model for 50 epochs, with a batch size of 16 and a learning rate of  $10^{-4}$.


For the evaluation of the trained agents, we provide both open-loop and closed-loop results. Open-loop results are calculated by computing the average displacement error (ADE), the final displacement error (FDE), the average head error (AHE), and the final head error (FHE) between the predicted and ground-truth trajectories as it is calculated in NuPlan~\cite{nuplan}. We provide our metrics for 3 different prediction horizons in seconds with 2Hz sampling rate. We used the Town13 validation set of our dataset. See Table~\ref{tab:openloop-results} for the results. For the closed-loop evaluation, we simplified our validation dataset routes to 36 scenarios and run them with the CARLA Leaderboard V2~\cite{CARLA} framework. We provide scenario-specific open-loop and closed-loop results in the Appendix. 

\subsection{Evaluation Results of Planning Agents}
\label{appendix planning evaluations}

Closed-loop metrics are summarized in Table~\ref{closed-loop-table}. These metric scores were obtained using leaderboard 2.0 original metrics in the challenge. The routes of the scenarios are randomly selected in Town13 scenarios in Leaderboard 2.0.  

\begin{table}[h]
    \centering
    \caption{Closed-loop metrics in unseen Town13 (Driving Score ↑, Route Score ↑, Penalty ↑) on validation routes. PlanT* is trained with the samples whose corresponding rarity scores are greater than 0. }
    \begin{tabular}{lccc}
        \toprule
        \textbf{Model} & \textbf{Driving} & \textbf{Route} & \textbf{Penalty} \\
        \midrule
        DiffusionDrive \cite{diffusiondrive}  & 22.35 & 62.06 & 0.339 \\
        Transfuser \cite{transfuser}     & 17.18 & 65.67 & 0.283 \\
        PlanT  \cite{planT}        & 52.95 & 81.67 & 0.658  \\
        PlanT* \cite{planT}        & 59.25 & 81.59 & 0.705  \\
        \bottomrule
    \end{tabular}
    \label{closed-loop-table}
\end{table}

\section{Conclusion}
\label{sec:conclusion}
We have developed a new dataset that integrates the strengths of previous efforts, including the robust PDM model and the versatile NuScenes \cite{caesar2020nuscenes} sensor configuration. This dataset is designed to address existing limitations, offering broad applicability across both end-to-end and modular systems. By incorporating a comprehensive set of tasks, ranging from dynamic object detection to traffic light recognition, our dataset aims to support cutting-edge research and improve model performance across various domains. As future work, since Bench2Drive and TaCarla share a similar sensor configuration but differ in their expert policies, training a planning model on the combined dataset could be a promising direction to mitigate the individual weaknesses of each expert and improve overall model robustness. 





\clearpage

{
    \small
    \bibliographystyle{ieeenat_fullname}
    \bibliography{main}
}

\clearpage
\clearpage
\setcounter{page}{1}

\section{Supplementary}
\begin{figure}[h]
  \centering
  \includegraphics[width=6cm]{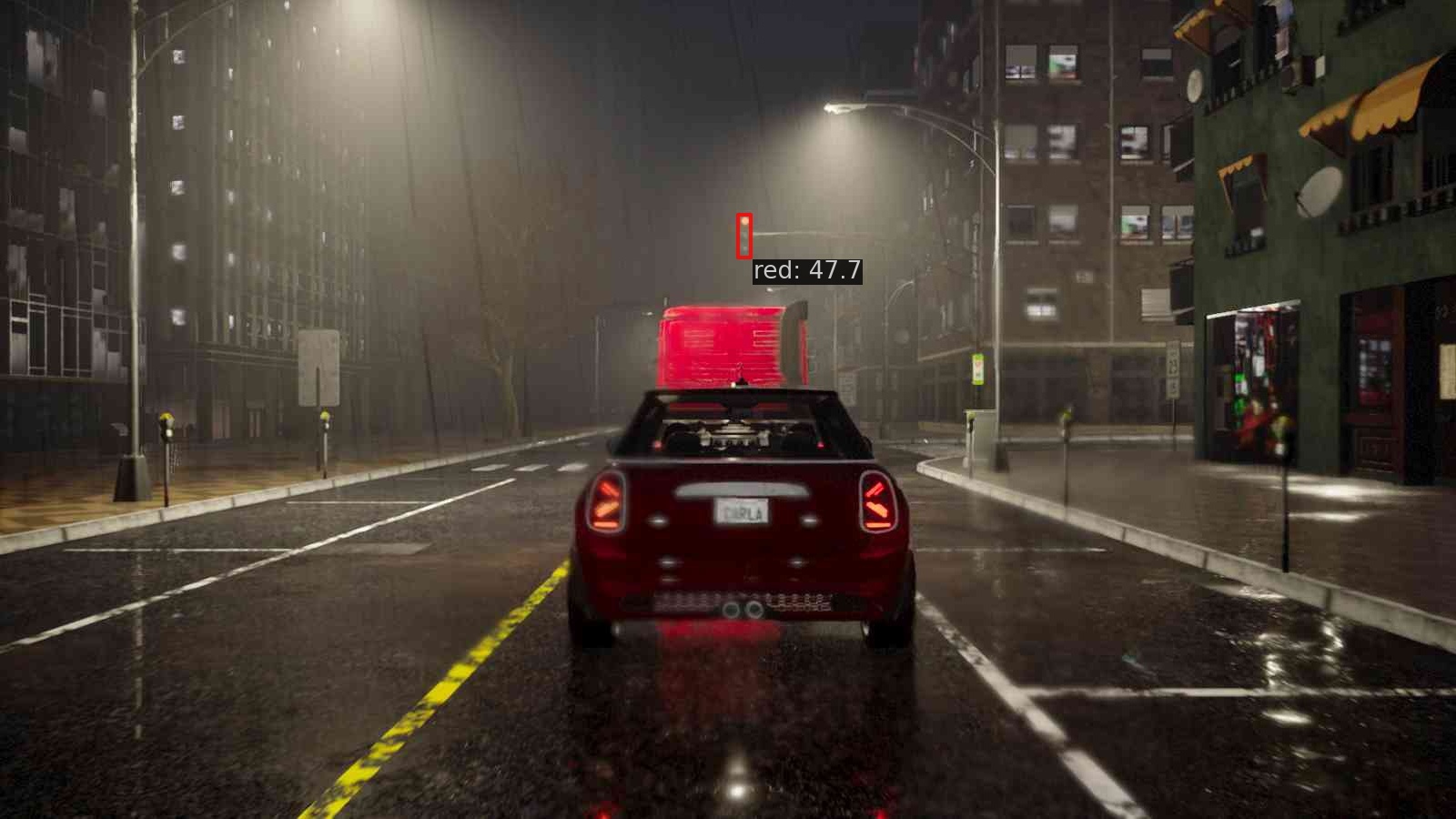} 
  \includegraphics[width=6cm]{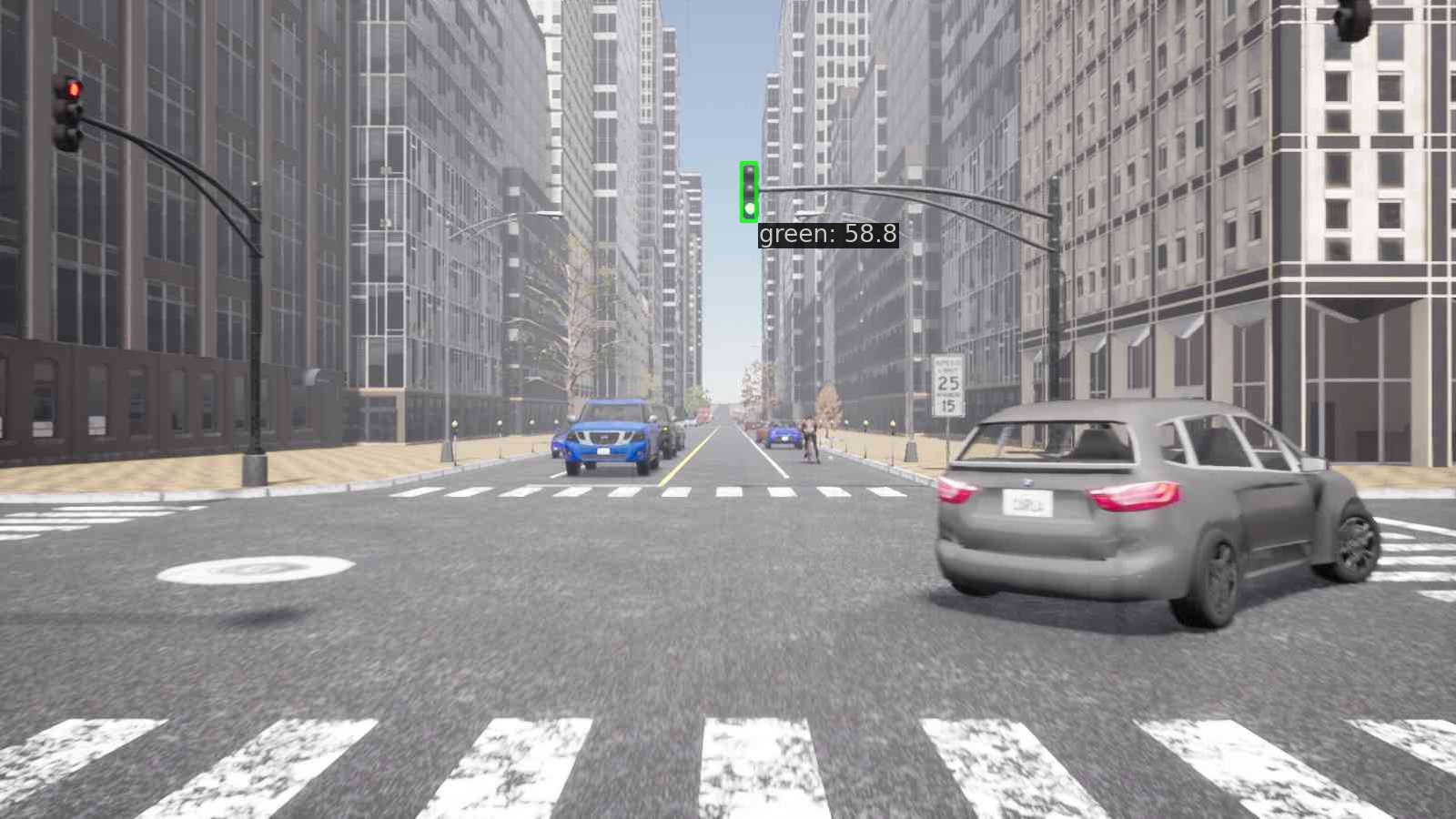}
  \caption{Outputs of the FCOS traffic light model}
\end{figure}

\subsection{Annotation Details}
\label{appendix annotation details}

Samples in TaCarla were collected using the PlanT~\cite{planT} data pipeline. An additional translation is applied to compensate for the virtual ego-vehicle center (1.3\,m in \(x\), 2.5\,m in \(z\)). All labels are stored as Parquet files, which can be read with PySpark to reduce label loading time. In addition, we provide code for reading labels from Parquet files, making it easier for others to use the dataset.

\begin{center}
  \includegraphics[width=0.48\textwidth]{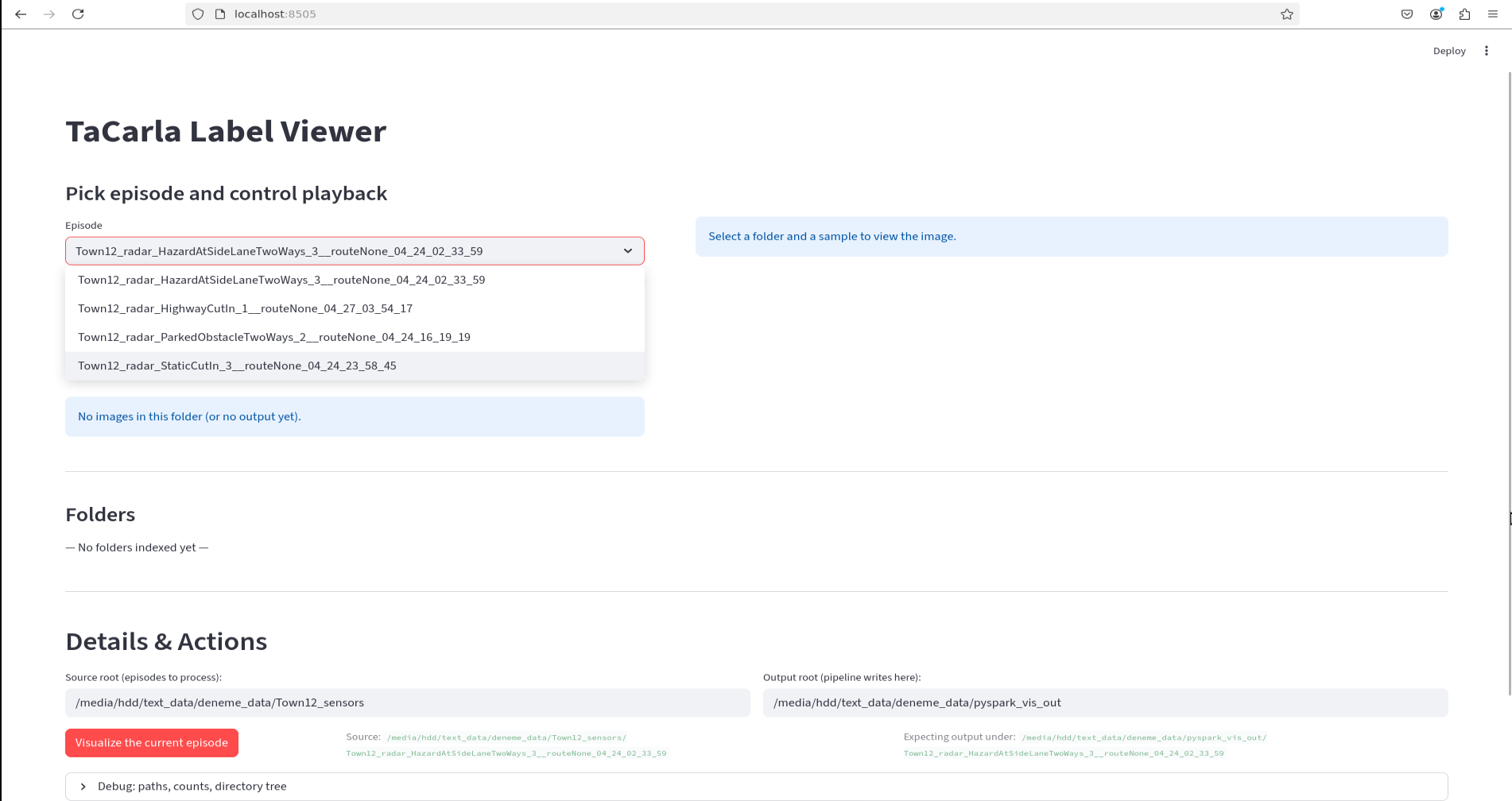}
\end{center}

\subsection{Scenarios}
\label{appendix scenarios}

The scenarios are arranged in the CARLA 9.15 simulator to increase diversity for learning-based planning. Unlike Leaderboard 1.0, Leaderboard 2.0 requires learning overtaking, stopping, merging, and avoiding dynamic actors. The following figures illustrate each scenario. Pink waypoints show future ego positions (8–30 frames ahead), warped to the current coordinate frame.
\FloatBarrier

\begin{figure}[t]
    \centering
    \begin{minipage}{0.48\linewidth}
        \includegraphics[width=\linewidth]{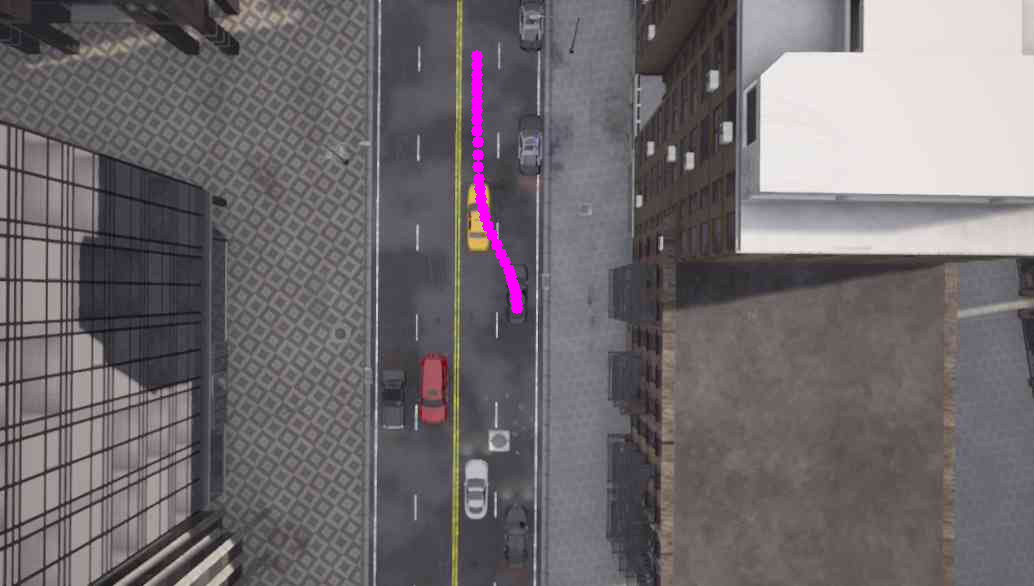}
        \caption*{\small \textbf{Accident}: Ego overtakes parked multiple vehicles.}
    \end{minipage}\hfill
    \begin{minipage}{0.48\linewidth}
        \includegraphics[width=\linewidth]{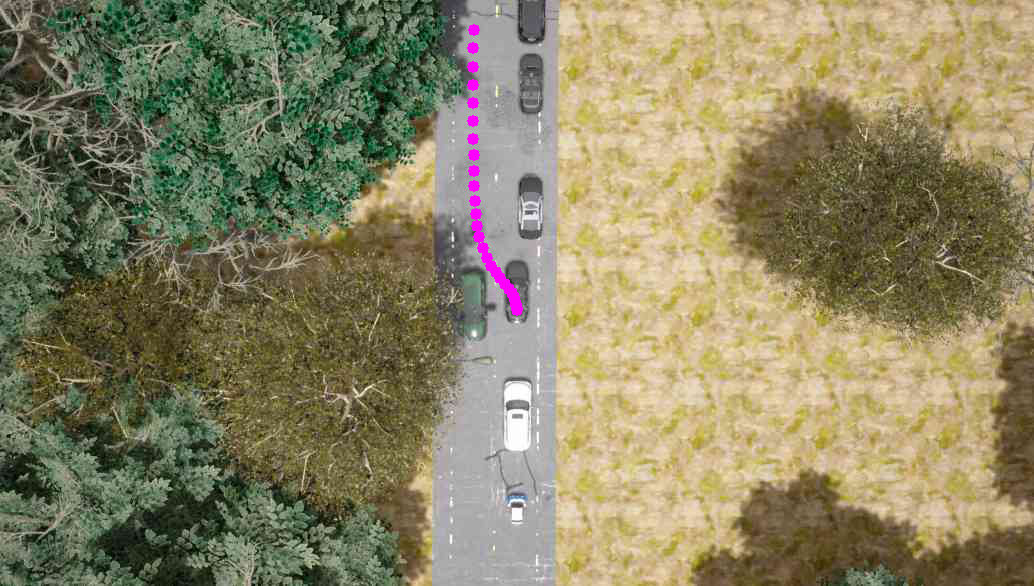}
        \caption*{\small \textbf{AccidentTwoWays}: Overtaking parked multiple vehicles into the opposite lane.}
    \end{minipage}
\end{figure}

\begin{figure}[t]
    \centering
    \begin{minipage}{0.45\linewidth}
        \includegraphics[width=\linewidth]{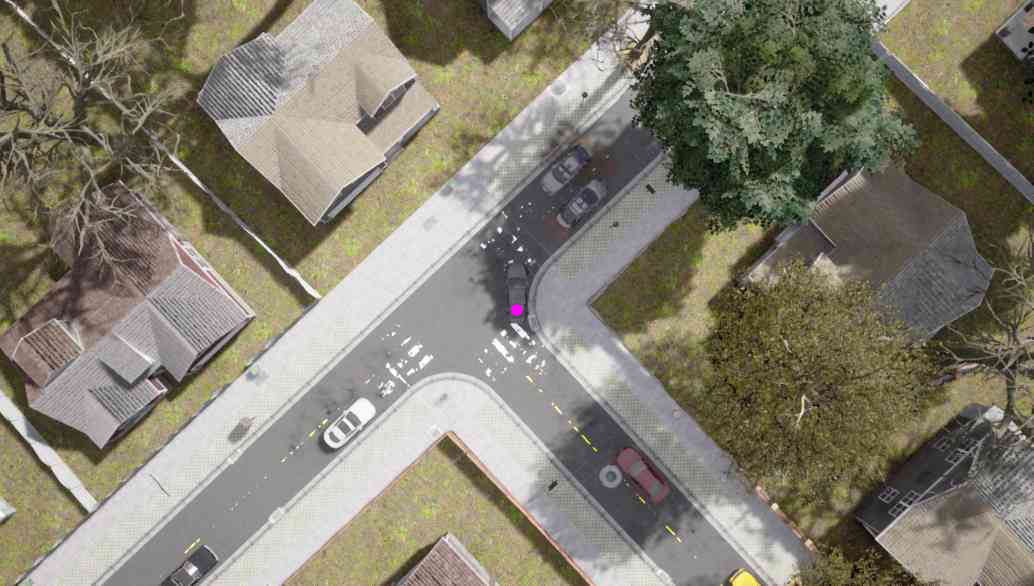}
        \caption*{\small \textbf{BlockedIntersection}: Intersection blocked.}
    \end{minipage}\hfill
    \begin{minipage}{0.45\linewidth}
        \includegraphics[width=\linewidth]{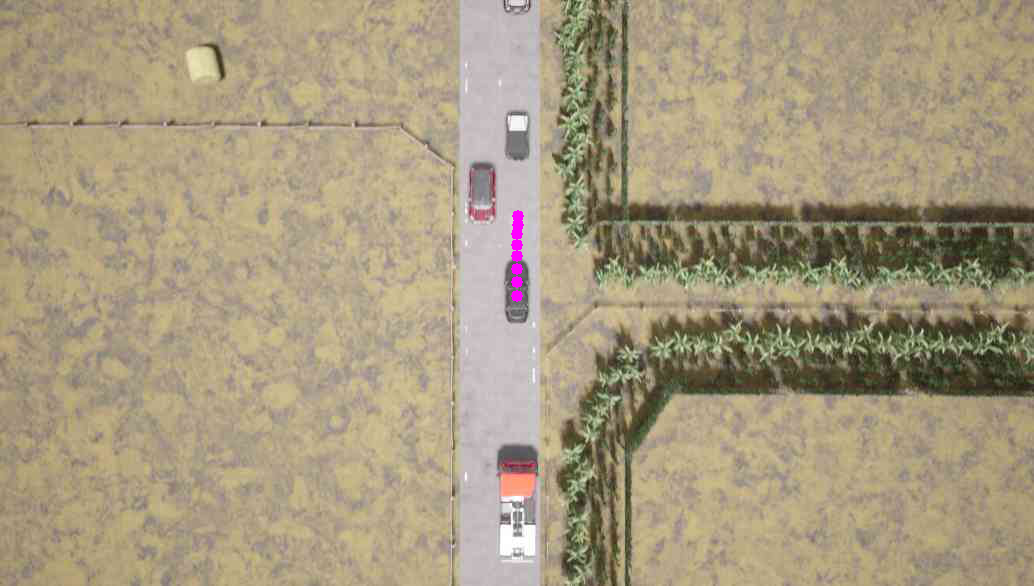}
        \caption*{\small \textbf{ControlLoss}: Temporary loss of vehicle control.}
    \end{minipage}
\end{figure}

\begin{figure}[t]
    \centering
    \begin{minipage}{0.48\linewidth}
        \includegraphics[width=\linewidth]{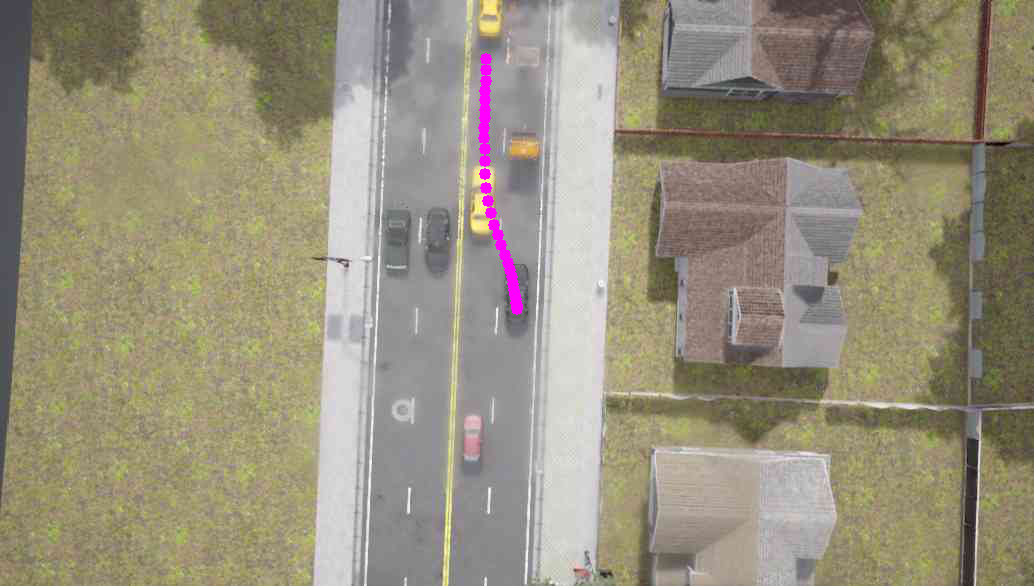}
        \caption*{\small \textbf{ConstructionObstacle}: Lane blocked by construction.}
    \end{minipage}\hfill
    \begin{minipage}{0.48\linewidth}
        \includegraphics[width=\linewidth]{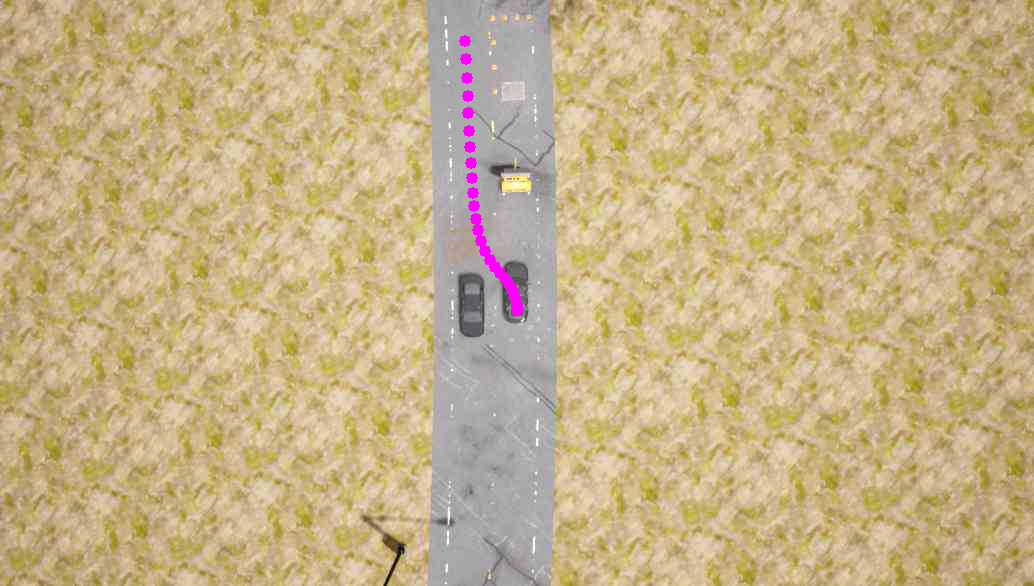}
        \caption*{\small \textbf{ConstructionObstacleTwoWays}: Overtaking via oncoming lane.}
    \end{minipage}
\end{figure}

\begin{figure}[t]
    \centering
    \begin{minipage}{0.48\linewidth}
        \includegraphics[width=\linewidth]{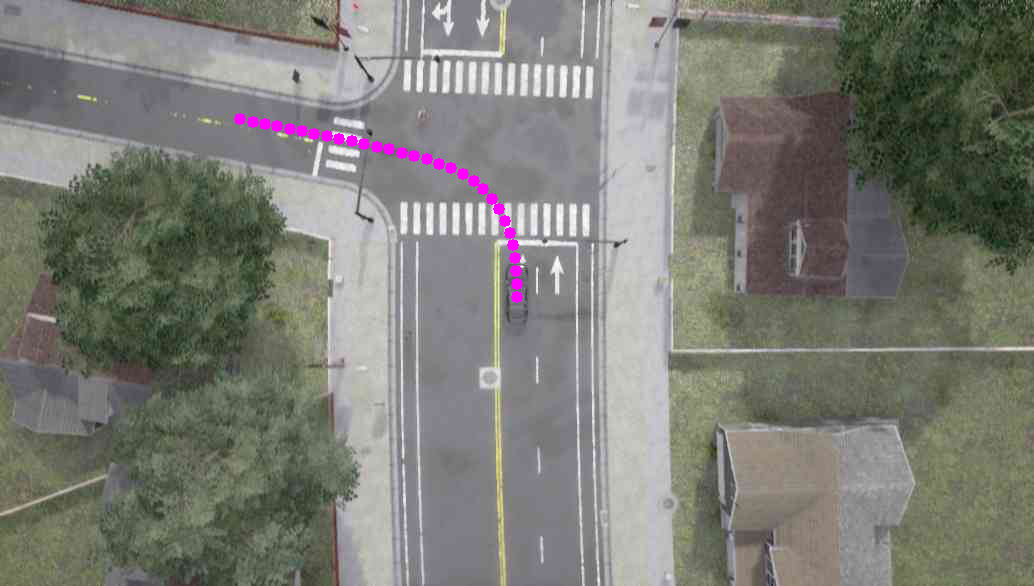}
        \caption*{\small \textbf{CrossingBicycleFlow}: Bicycles crossing the junction.}
    \end{minipage}\hfill
    \begin{minipage}{0.48\linewidth}
        \includegraphics[width=\linewidth]{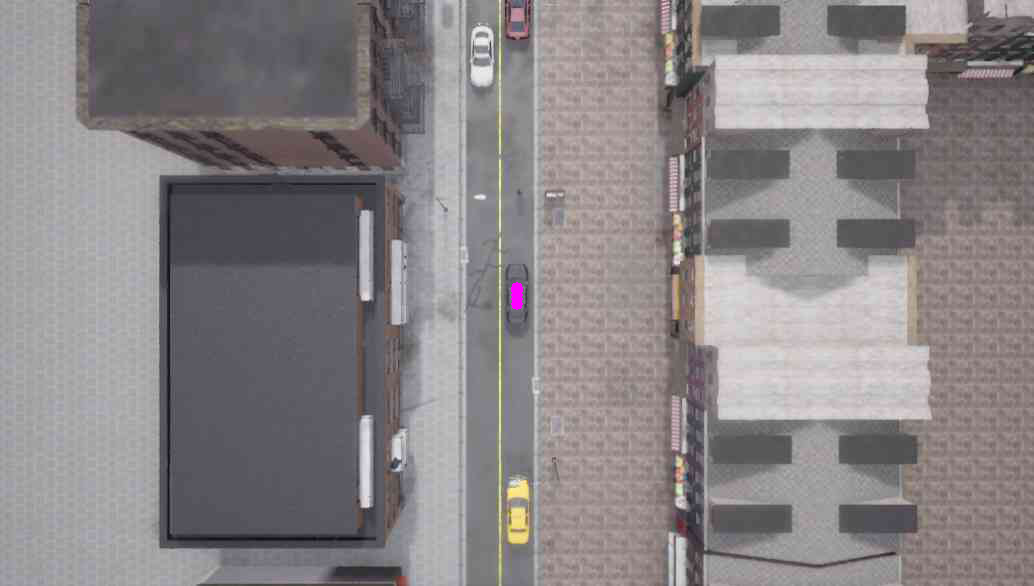}
        \caption*{\small \textbf{DynamicObjectCrossing} / \textbf{PedestrianCrossing}: Pedestrian or cyclist crossing.}
    \end{minipage}
\end{figure}

\begin{figure}[t]
    \centering
    \begin{minipage}{0.48\linewidth}
        \includegraphics[width=\linewidth]{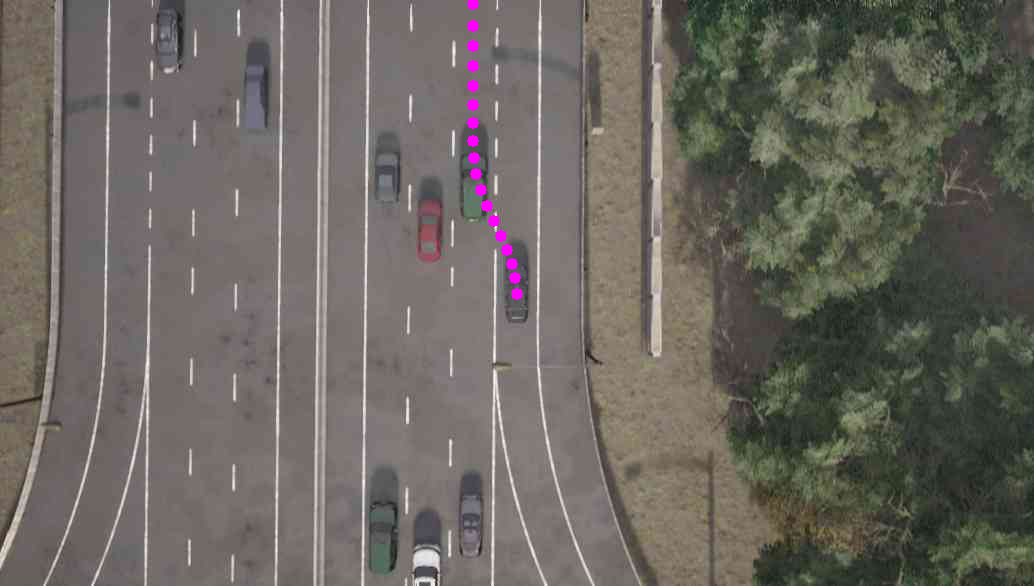}
        \caption*{\small \textbf{EnterActorFlow}/\textbf{V2}: Joining a main road with traffic flow.}
    \end{minipage}\hfill
    \begin{minipage}{0.48\linewidth}
        \includegraphics[width=\linewidth]{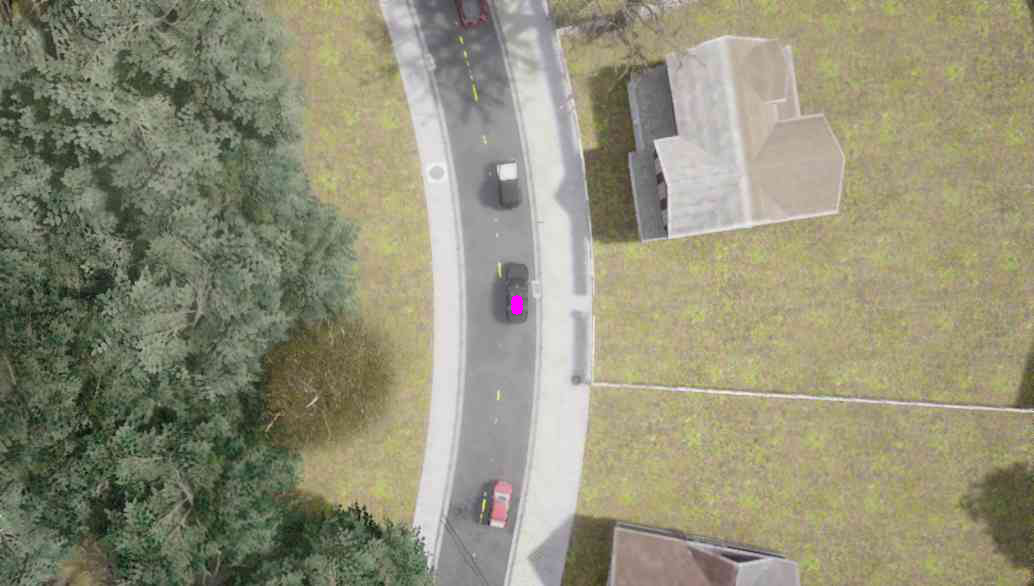}
        \caption*{\small \textbf{HardBrakeRoute}: Sudden stop of lead vehicle.}
    \end{minipage}
\end{figure}

\begin{figure}[t]
    \centering
    \begin{minipage}{0.48\linewidth}
        \includegraphics[width=\linewidth]{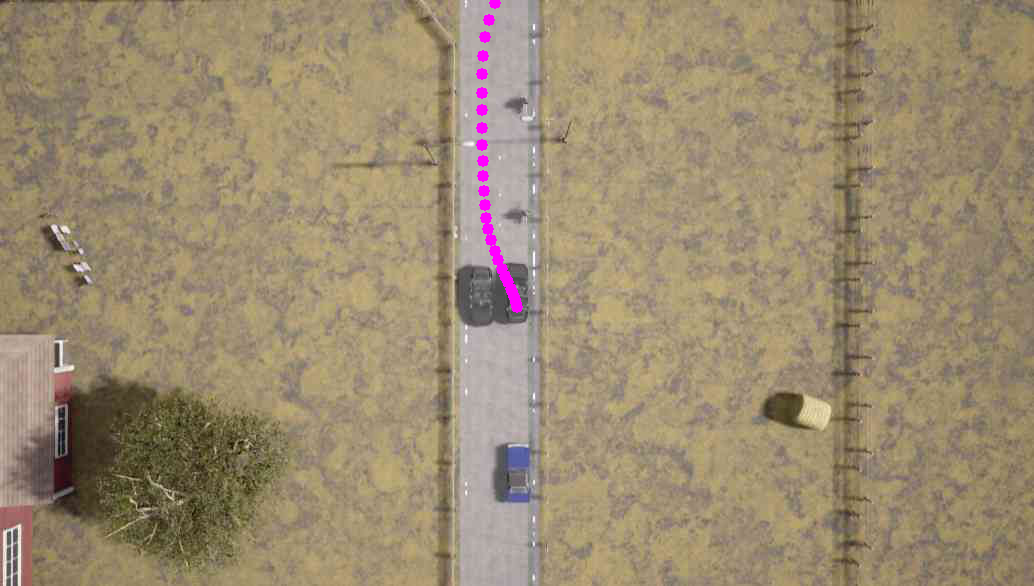}
        \caption*{\small \textbf{HazardAtSideLane}: Slow bicycles on roadside.}
    \end{minipage}\hfill
    \begin{minipage}{0.48\linewidth}
        \includegraphics[width=\linewidth]{scenario_images/HazardAtSideLaneTwoWays/HazardAtSideLaneTwoWays.png}
        \caption*{\small \textbf{HazardAtSideLaneTwoWays}: Overtaking bicycles via opposite lane.}
    \end{minipage}
\end{figure}

\begin{figure}[t]
    \centering
    \begin{minipage}{0.48\linewidth}
        \includegraphics[width=\linewidth]{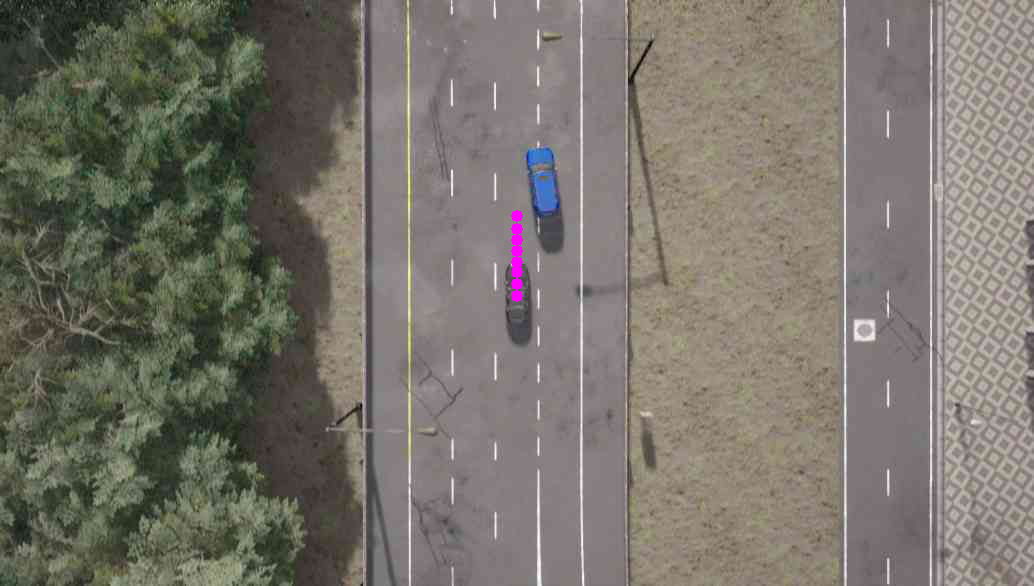}
        \caption*{\small \textbf{HighwayCutIn}: Another vehicle cuts in.}
    \end{minipage}\hfill
    \begin{minipage}{0.48\linewidth}
        \includegraphics[width=\linewidth]{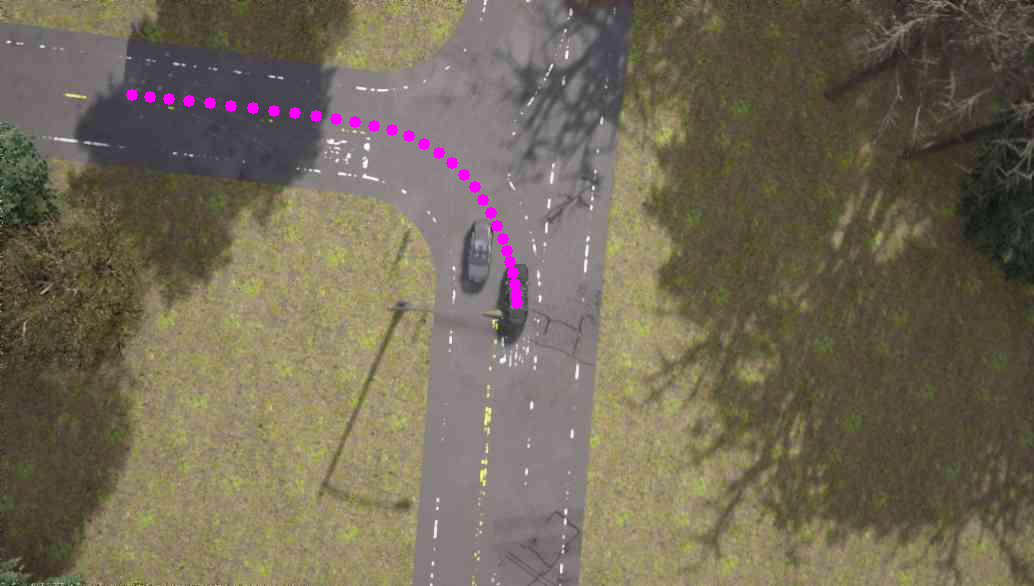}
        \caption*{\small \textbf{InterurbanActorFlow}/\textbf{Advanced}: Dense intersection flow.}
    \end{minipage}
\end{figure}

\begin{figure}[t]
    \centering
    \begin{minipage}{0.48\linewidth}
        \includegraphics[width=\linewidth]{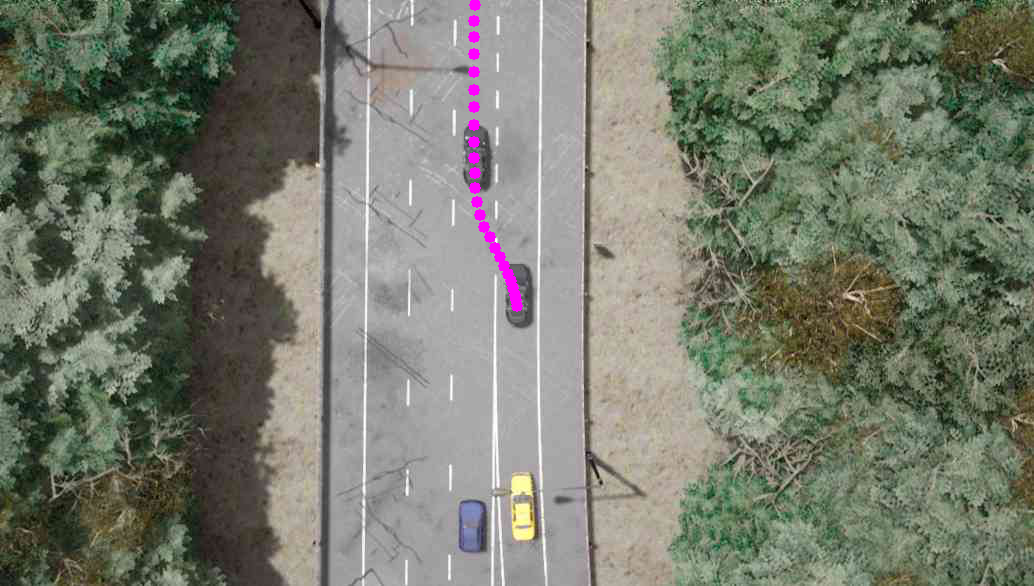}
        \caption*{\small \textbf{MergingIntoSlowTraffic}: Merging onto slow-moving road.}
    \end{minipage}\hfill
    \begin{minipage}{0.48\linewidth}
        \includegraphics[width=\linewidth]{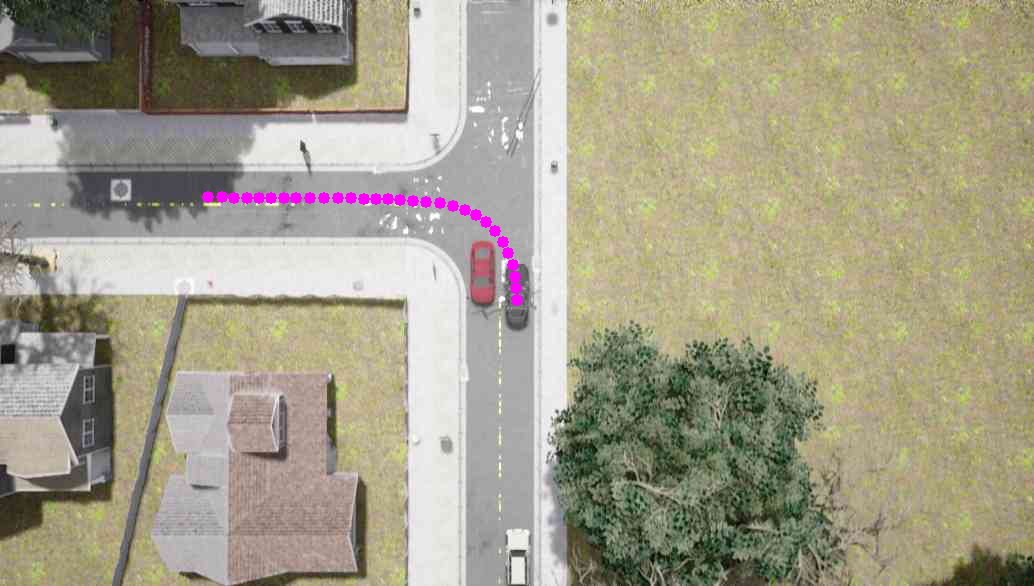}
        \caption*{\small \textbf{NonSignalizedJunctionLeft/RightTurn}: Turning without traffic lights.}
    \end{minipage}
\end{figure}

\begin{figure}[t]
    \centering
    \begin{minipage}{0.48\linewidth}
        \includegraphics[width=\linewidth]{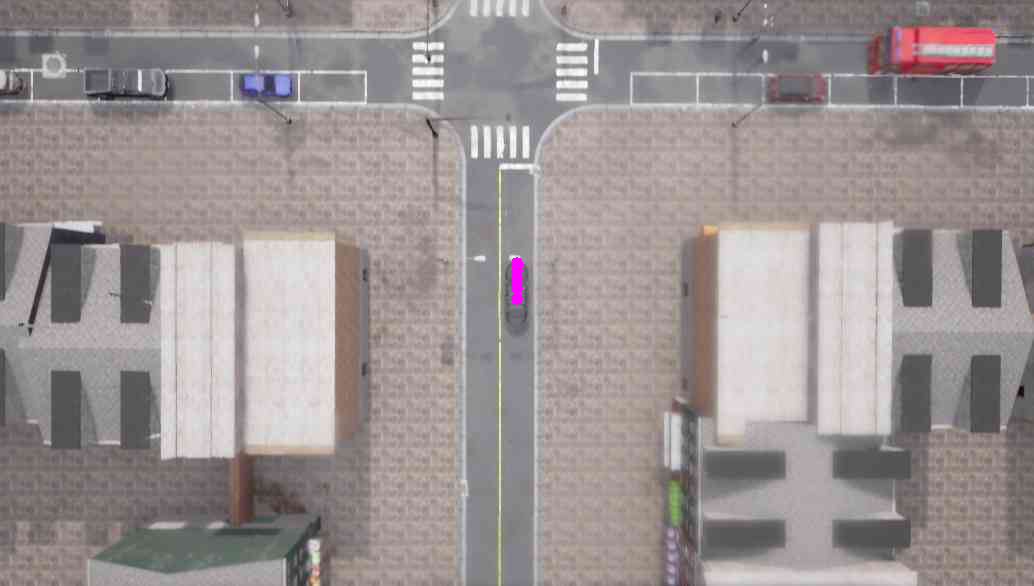}
        \caption*{\small \textbf{OppositeVehicleRunningRedLight/TakingPriority}.}
    \end{minipage}\hfill
    \begin{minipage}{0.48\linewidth}
        \includegraphics[width=\linewidth]{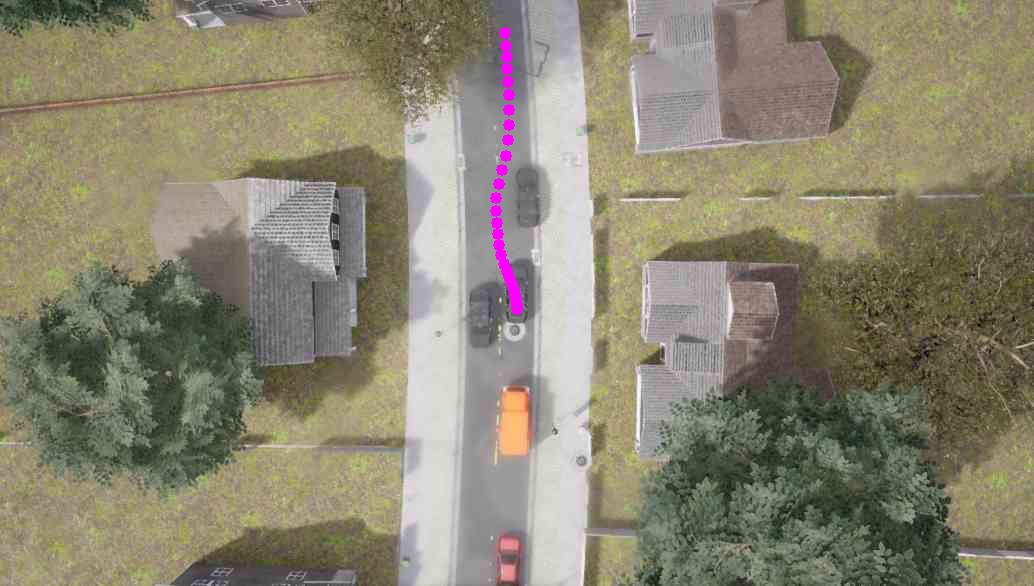}
        \caption*{\small \textbf{ParkedObstacle}/\textbf{TwoWays}: Overtaking a parked vehicle.}
    \end{minipage}
\end{figure}

\begin{figure}[t]
    \centering
    \begin{minipage}{0.48\linewidth}
        \includegraphics[width=\linewidth]{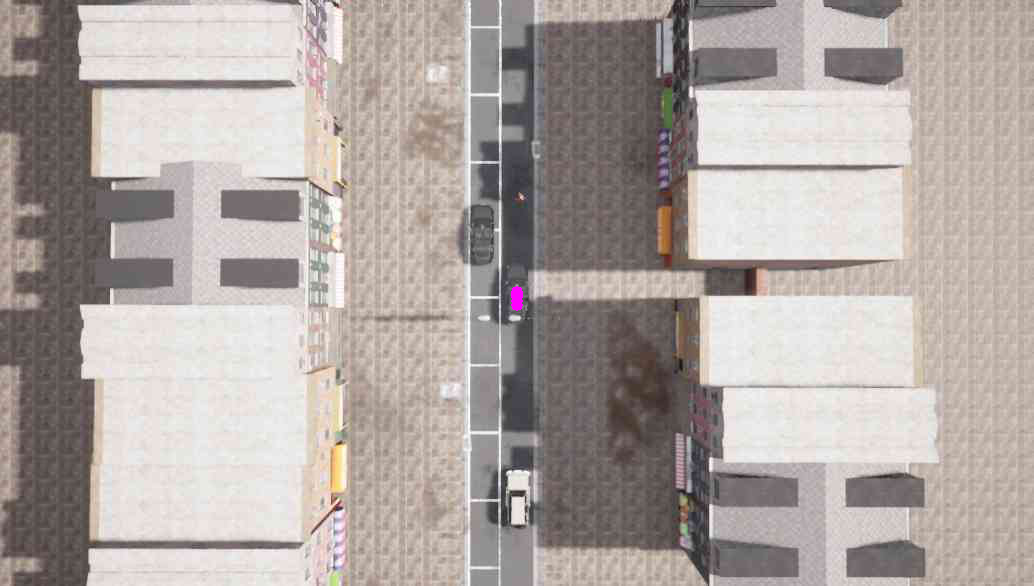}
        \caption*{\small \textbf{ParkingCrossingPedestrian}: Pedestrian emerging from behind a vehicle.}
    \end{minipage}\hfill
    \begin{minipage}{0.48\linewidth}
        \includegraphics[width=\linewidth]{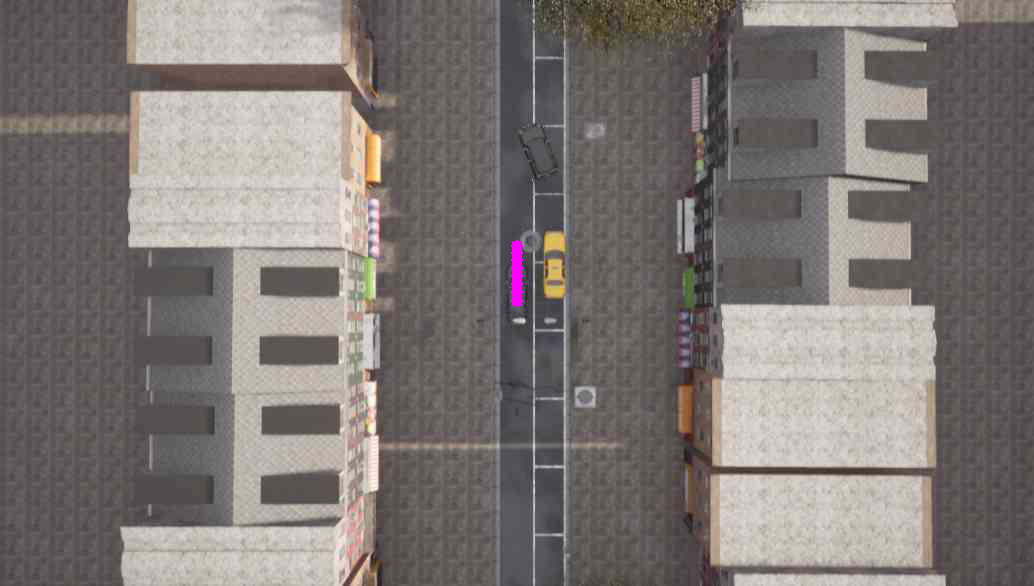}
        \caption*{\small \textbf{ParkingCutIn}: Parked vehicle suddenly merges.}
    \end{minipage}
\end{figure}

\begin{figure}[t]
    \centering
    \begin{minipage}{0.48\linewidth}
        \includegraphics[width=\linewidth]{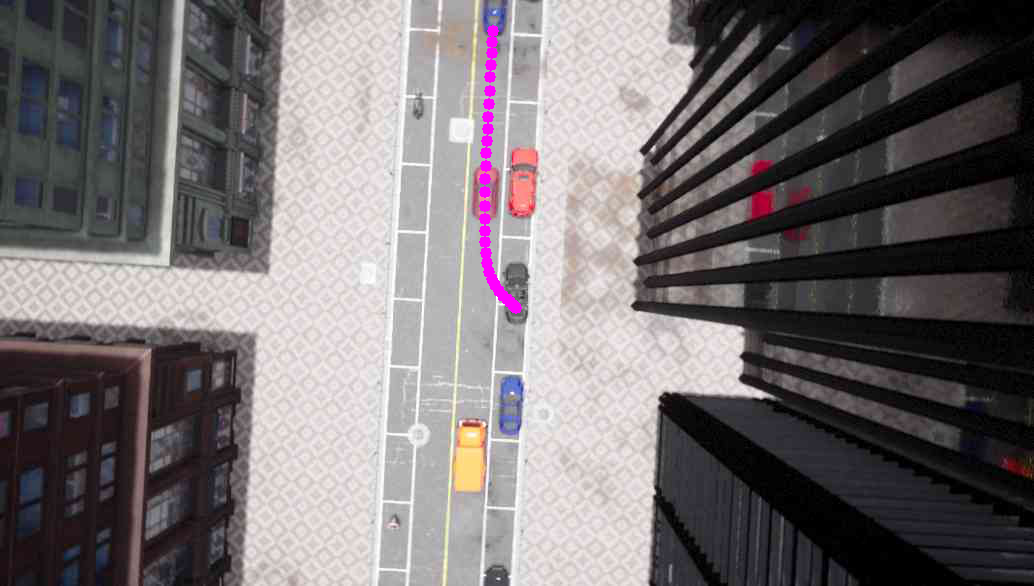}
        \caption*{\small \textbf{ParkingExit}: Ego exits a parking spot.}
    \end{minipage}\hfill
    \begin{minipage}{0.48\linewidth}
        \includegraphics[width=\linewidth]{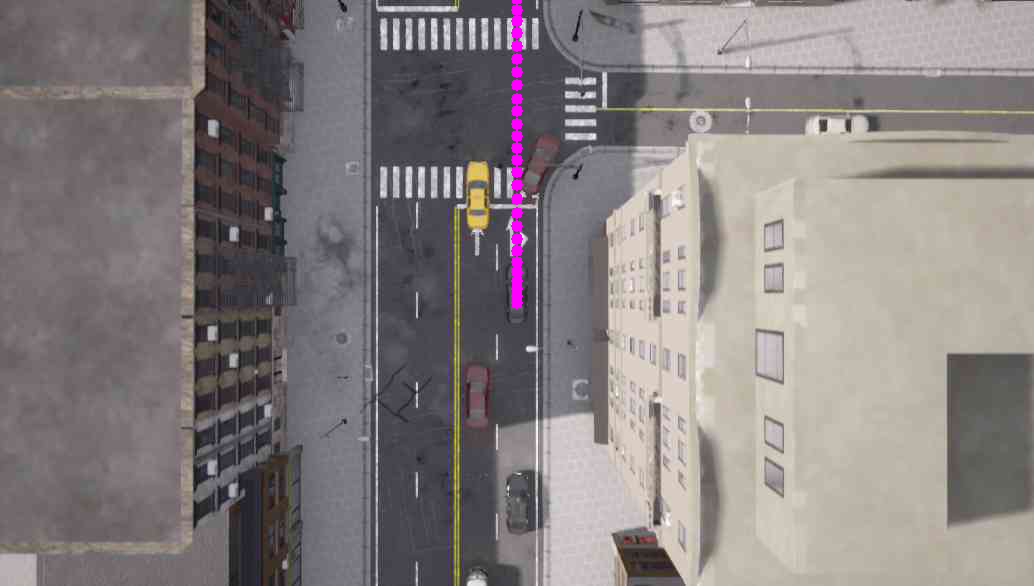}
        \caption*{\small \textbf{PriorityAtJunction}: Ego is prioritized to pass.}
    \end{minipage}
\end{figure}

\begin{figure}[t]
    \centering
    \begin{minipage}{0.48\linewidth}
        \includegraphics[width=\linewidth]{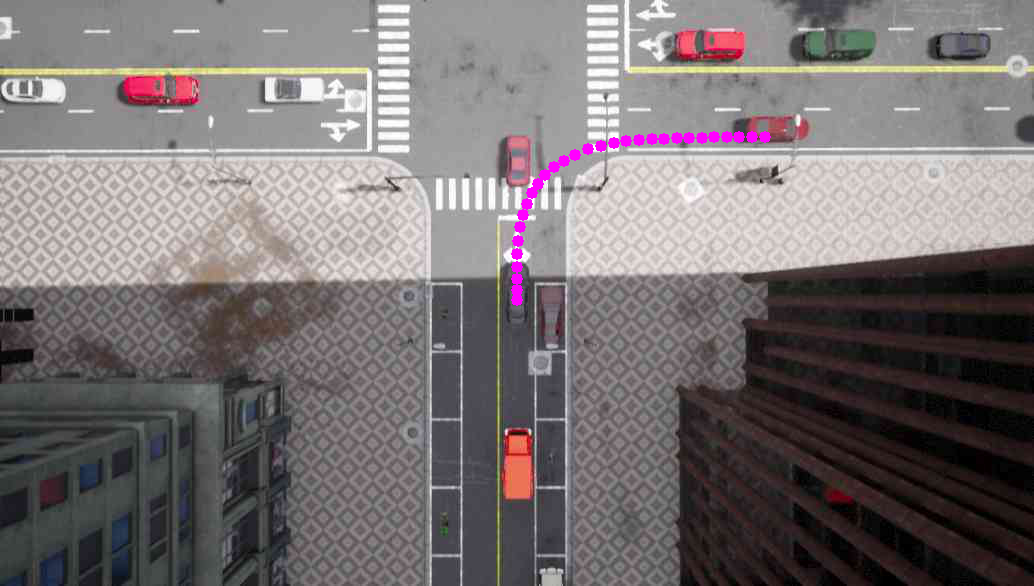}
        \caption*{\small \textbf{SignalizedJunctionLeft/RightTurn}: Turns at signalized junction.}
    \end{minipage}\hfill
    \begin{minipage}{0.48\linewidth}
        \includegraphics[width=\linewidth]{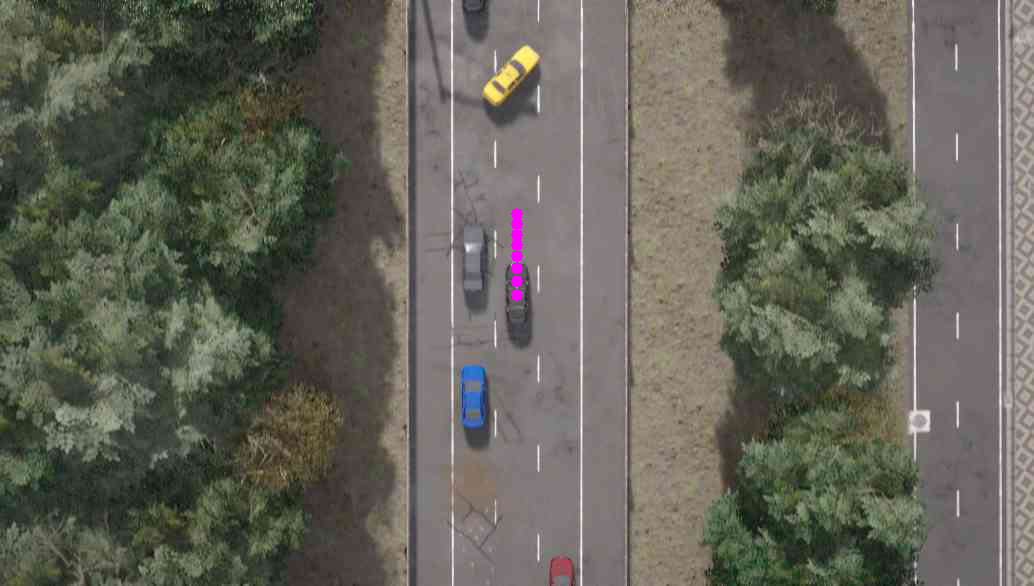}
        \caption*{\small \textbf{StaticCutIn}: Stationary vehicle suddenly cuts in.}
    \end{minipage}
\end{figure}

\begin{figure}[t]
    \centering
    \begin{minipage}{0.48\linewidth}
        \includegraphics[width=\linewidth]{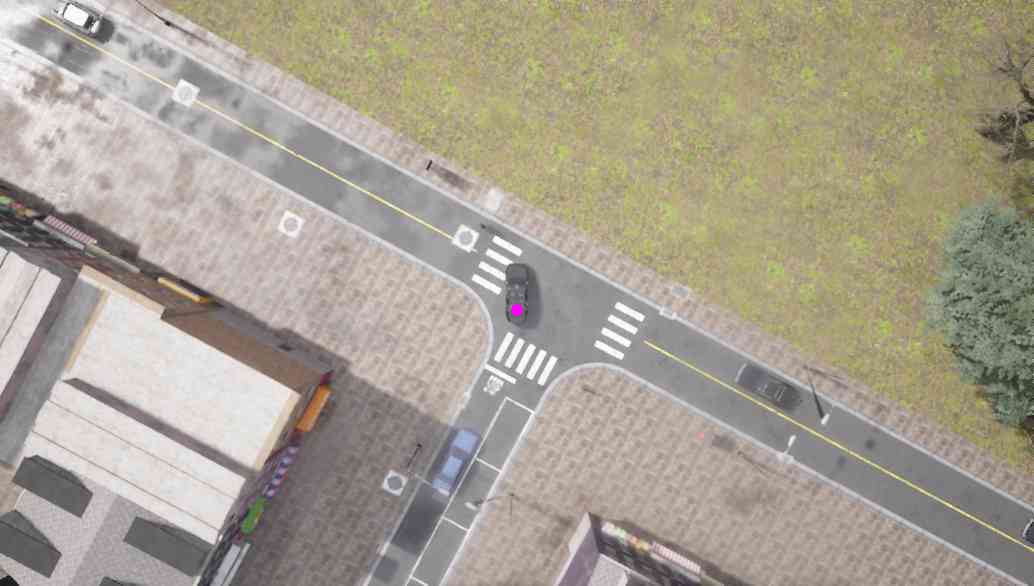}
        \caption*{\small \textbf{VehicleTurningRoute/ Pedestrian}: Pedestrian during turning.}
    \end{minipage}\hfill
    \begin{minipage}{0.48\linewidth}
        \includegraphics[width=\linewidth]{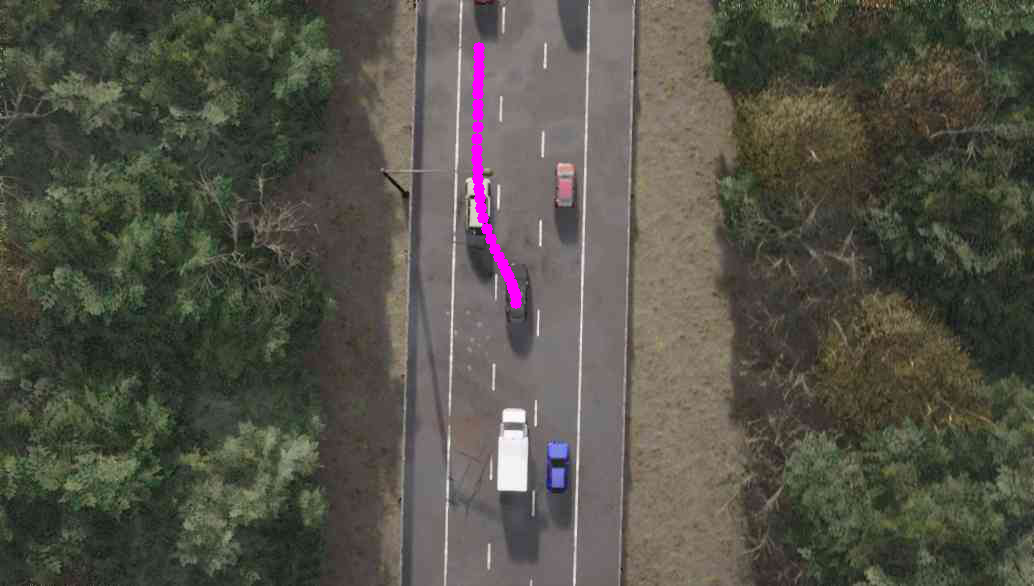}
        \caption*{\small \textbf{YieldToEmergencyVehicle}: Ego yields to emergency vehicle.}
    \end{minipage}
\end{figure}

\clearpage

\begin{table*}[t]
  \centering
  \scriptsize
  \setlength{\tabcolsep}{3pt}
  \caption{Open-loop metrics (ADE ↓, FDE ↓, AHE ↓, FHE ↓) for each scenario.}
  \label{open-loop-table-full}

  \begin{minipage}[t]{0.45\textwidth}
    \centering
    \begin{tabular}{lccccc}
      \toprule
      \textbf{Scenario Name} & \textbf{Model} & ADE ↓ & FDE ↓ & AHE ↓ & FHE ↓ \\
      \midrule

      \multirow{3}{*}{Accident}
        & DiffusionDrive & 2.1493 & 4.0263 & 0.1040 & 0.1156 \\
        & PlanT          & 1.4002 & 2.2346 & 0.1368 & 0.1934 \\
        & Transfuser     & 1.8716 & 3.4446 & 0.1109 & 0.0963 \\
      \midrule

      \multirow{3}{*}{AccidentTwoWays}
        & DiffusionDrive & 0.8104 & 1.5032 & 0.6924 & 0.6760 \\
        & PlanT          & 0.5810 & 2.2346 & 0.1368 & 0.1934 \\
        & Transfuser     & 0.5422 & 1.0367 & 0.8329 & 1.0033 \\
      \midrule

      \multirow{3}{*}{BlockedIntersection}
        & DiffusionDrive & 1.0451 & 1.9725 & 0.4265 & 0.4387 \\
        & PlanT          & 1.2397 & 2.0076 & 0.6148 & 0.8359 \\
        & Transfuser     & 1.1008 & 2.1835 & 0.4541 & 0.5256 \\
      \midrule

      \multirow{3}{*}{ConstructionObstacleTwoWays}
        & DiffusionDrive & 0.7603 & 1.4043 & 0.6028 & 0.6724 \\
        & PlanT          & 2.8932 & 4.9572 & 0.0632 & 0.0806 \\
        & Transfuser     & 0.4308 & 0.7822 & 0.7733 & 1.0419 \\
      \midrule

      \multirow{3}{*}{ControlLoss}
        & DiffusionDrive & 0.8896 & 1.6741 & 0.1025 & 0.1024 \\
        & PlanT          & 0.5790 & 0.9722 & 0.1924 & 0.2077 \\
        & Transfuser     & 0.5935 & 1.1126 & 0.0787 & 0.0853 \\
      \midrule

      \multirow{3}{*}{DynamicObjectCrossing}
        & DiffusionDrive & 0.9871 & 1.8262 & 0.1476 & 0.1331 \\
        & PlanT          & 0.9846 & 1.5602 & 0.3478 & 0.3785 \\
        & Transfuser     & 0.8131 & 1.5747 & 0.1357 & 0.1465 \\
      \midrule

      \multirow{3}{*}{EnterActorFlow}
        & DiffusionDrive & 1.1639 & 2.2517 & 0.0301 & 0.0411 \\
        & PlanT          & 1.6695 & 2.7945 & 0.0224 & 0.0246 \\
        & Transfuser     & 1.2159 & 2.2782 & 0.0302 & 0.0357 \\
      \midrule

      \multirow{3}{*}{HardBreakRoute}
        & DiffusionDrive & 0.9202 & 1.7693 & 0.3500 & 0.3283 \\
        & PlanT          & 0.3669 & 0.6509 & 0.5425 & 0.4152 \\
        & Transfuser     & 0.8418 & 1.6670 & 0.3553 & 0.2879 \\
      \midrule

      \multirow{3}{*}{HazardAtSideLane}
        & DiffusionDrive & 1.8221 & 3.4470 & 0.0961 & 0.1010 \\
        & PlanT          & 2.3169 & 3.7521 & 0.1863 & 0.1486 \\
        & Transfuser     & 1.8582 & 3.5167 & 0.0875 & 0.0902 \\
      \midrule

      \multirow{3}{*}{HazardAtSideLaneTwoWays}
        & DiffusionDrive & 1.2479 & 2.4540 & 0.0458 & 0.0490 \\
        & PlanT          & 1.8753 & 3.2281 & 0.0328 & 0.0347 \\
        & Transfuser     & 0.9637 & 1.8608 & 0.0225 & 0.0294 \\
      \midrule

      \multirow{3}{*}{HighwayExit}
        & DiffusionDrive & 1.3126 & 2.5442 & 0.0373 & 0.0422 \\
        & PlanT          & 1.1354 & 1.9507 & 0.0620 & 0.0709 \\
        & Transfuser     & 1.0215 & 1.9594 & 0.0439 & 0.0355 \\
      \midrule

      \multirow{3}{*}{MergerIntoSlowTraffic}
        & DiffusionDrive & 2.4493 & 4.5116 & 0.0762 & 0.0851 \\
        & PlanT          & 3.1610 & 5.2284 & 0.1672 & 0.1563 \\
        & Transfuser     & 1.8243 & 3.3578 & 0.0638 & 0.0779 \\
      \midrule

      \multirow{3}{*}{MergerIntoSlowTrafficV2}
        & DiffusionDrive & 1.9542 & 3.6690 & 0.1677 & 0.1683 \\
        & PlanT          & 1.8335 & 3.1872 & 0.1761 & 0.1634 \\
        & Transfuser     & 1.9213 & 3.8040 & 0.1369 & 0.1280 \\
      \midrule

      \multirow{3}{*}{NonSignalizedJunctionLeftTurn}
        & DiffusionDrive & 1.3052 & 2.4879 & 0.1387 & 0.1602 \\
        & PlanT          & 1.4098 & 2.3628 & 0.1842 & 0.2272 \\
        & Transfuser     & 1.1051 & 2.1444 & 0.1293 & 0.1526 \\
      \midrule

      \multirow{3}{*}{NonSignalizedJunctionRightTurn}
        & DiffusionDrive & 1.6663 & 3.3112 & 0.1150 & 0.1305 \\
        & PlanT          & 1.8390 & 3.1044 & 0.2153 & 0.2461 \\
        & Transfuser     & 1.3446 & 2.6837 & 0.1063 & 0.1106 \\
      \bottomrule
    \end{tabular}
  \end{minipage}
  \hfill
  \begin{minipage}[t]{0.48\textwidth}
    \centering
    \begin{tabular}{lccccc}
      \toprule
      \textbf{Scenario Name} & \textbf{Model} & ADE ↓ & FDE ↓ & AHE ↓ & FHE ↓ \\
      \midrule

      \multirow{3}{*}{OppositeVehicleRunningRedlight}
        & DiffusionDrive & 1.6435 & 3.1160 & 0.0324 & 0.0364 \\
        & PlanT          & 1.3400 & 2.1328 & 0.0350 & 0.0178 \\
        & Transfuser     & 1.3820 & 2.6198 & 0.0085 & 0.0074 \\
      \midrule

      \multirow{3}{*}{OppositeVehicleTakingPriority}
        & DiffusionDrive & 0.8963 & 1.6087 & 0.0521 & 0.0491 \\
        & PlanT          & 1.7681 & 2.9627 & 0.0265 & 0.0231 \\
        & Transfuser     & 0.8336 & 1.5672 & 0.0156 & 0.0157 \\
      \midrule

      \multirow{3}{*}{ParkedObstacle}
        & DiffusionDrive & 3.8699 & 7.3543 & 0.0734 & 0.0995 \\
        & PlanT          & 1.6408 & 2.6752 & 0.2131 & 0.1079 \\
        & Transfuser     & 2.9464 & 5.6412 & 0.0278 & 0.0309 \\
      \midrule

      \multirow{3}{*}{ParkedObstacleTwoWays}
        & DiffusionDrive & 0.8414 & 1.5375 & 0.2680 & 0.2518 \\
        & PlanT          & 0.7499 & 1.2404 & 0.2798 & 0.2334 \\
        & Transfuser     & 0.6938 & 1.2970 & 0.2149 & 0.2156 \\
      \midrule

      \multirow{3}{*}{ParkingCrossingPedestrian}
        & DiffusionDrive & 0.7730 & 1.3273 & 0.3226 & 0.3002 \\
        & PlanT          & 0.7523 & 1.1728 & 0.5261 & 0.4350 \\
        & Transfuser     & 0.5847 & 1.0419 & 0.4454 & 0.6068 \\
      \midrule

      \multirow{3}{*}{ParkingCutIn}
        & DiffusionDrive & 0.8182 & 1.4240 & 0.4070 & 0.3682 \\
        & PlanT          & 0.6404 & 0.9740 & 0.6235 & 0.5215 \\
        & Transfuser     & 0.8151 & 1.4573 & 0.4517 & 0.5596 \\
      \midrule

      \multirow{3}{*}{ParkingExit}
        & DiffusionDrive & 1.4377 & 2.7670 & 0.2318 & 0.2468 \\
        & PlanT          & 0.5360 & 0.9217 & 0.5211 & 0.2358 \\
        & Transfuser     & 1.0385 & 1.9858 & 0.1538 & 0.1154 \\
      \midrule

      \multirow{3}{*}{PedestrianCrossing}
        & DiffusionDrive & 1.8471 & 3.7742 & 0.1810 & 0.1688 \\
        & PlanT          & 1.7926 & 3.0981 & 0.6149 & 0.4200 \\
        & Transfuser     & 1.2441 & 2.5625 & 0.3430 & 0.3743 \\
      \midrule

      \multirow{3}{*}{PriorityAtJunction}
        & DiffusionDrive & 0.6663 & 1.1983 & 0.1608 & 0.1718 \\
        & PlanT          & 0.4824 & 0.7867 & 0.2022 & 0.2208 \\
        & Transfuser     & 0.6363 & 1.1397 & 0.2277 & 0.2609 \\
      \midrule

      \multirow{3}{*}{SignalizedJunctionLeftTurn}
        & DiffusionDrive & 2.1807 & 4.2420 & 0.2154 & 0.2378 \\
        & PlanT          & 2.1885 & 3.6065 & 0.2567 & 0.2515 \\
        & Transfuser     & 1.5983 & 3.1293 & 0.1678 & 0.2216 \\
      \midrule

      \multirow{3}{*}{SignalizedJunctionRightTurn}
        & DiffusionDrive & 1.7408 & 3.2851 & 0.1120 & 0.1567 \\
        & PlanT          & 1.6994 & 2.8513 & 0.1602 & 0.1629 \\
        & Transfuser     & 1.5451 & 2.9369 & 0.1474 & 0.1570 \\
      \midrule

      \multirow{3}{*}{StaticCutIn}
        & DiffusionDrive & 1.7001 & 3.2323 & 0.1431 & 0.1204 \\
        & PlanT          & 0.9219 & 1.5490 & 0.3423 & 0.3070 \\
        & Transfuser     & 1.1787 & 2.2042 & 0.1716 & 0.2178 \\
      \midrule

      \multirow{3}{*}{VehicleTurningRoute}
        & DiffusionDrive & 1.0698 & 1.9596 & 0.2788 & 0.2660 \\
        & PlanT          & 1.0357 & 1.7030 & 0.5682 & 0.6189 \\
        & Transfuser     & 0.9895 & 1.8191 & 0.3002 & 0.4327 \\
      \midrule

      \multirow{3}{*}{VehicleTurningRoutePedestrian}
        & DiffusionDrive & 1.0859 & 2.0227 & 0.2285 & 0.2530 \\
        & PlanT          & 0.8221 & 1.3853 & 0.6458 & 0.7563 \\
        & Transfuser     & 0.8494 & 1.6047 & 0.2154 & 0.2058 \\
      \midrule

      \multirow{3}{*}{YieldToEmergencyVehicle}
        & DiffusionDrive & 1.4413 & 3.0138 & 0.0252 & 0.0244 \\
        & PlanT          & 1.5263 & 2.5886 & 0.0889 & 0.1030 \\
        & Transfuser     & 0.6925 & 1.3326 & 0.0164 & 0.0214 \\
      \bottomrule
    \end{tabular}
  \end{minipage}
\end{table*}

\end{document}